\documentclass[journal]{IEEEtran}
\usepackage{lipsum} 
\usepackage{color}
\usepackage{graphicx}
\usepackage{amsmath}
\usepackage{amssymb}
\usepackage{epsfig}
\usepackage{array}
\usepackage{epstopdf}
\usepackage{amsthm}
\usepackage{afterpage}
\usepackage{amsmath,amssymb,amsfonts}
\usepackage{verbatim}
\usepackage{psfrag}
\usepackage{color}
\usepackage{cite}
\usepackage{longtable}
\usepackage{multirow}
\usepackage{setspace}
\usepackage{adjustbox}
\usepackage{epsfig}
\usepackage{epstopdf}
\usepackage{footmisc}
\usepackage{fmtcount}
\usepackage{stackrel}
\usepackage{float}
\usepackage{breqn}
\usepackage{enumerate}
\usepackage{graphicx}
\usepackage{subcaption}
\usepackage{tabularray}
\usepackage{algorithm}
\usepackage{algpseudocode}
\usepackage{dblfloatfix}
\usepackage{flushend}
\usepackage{graphicx}
\usepackage{dblfloatfix}

\usepackage{tabularx,booktabs}
\usepackage[labelformat=simple]{subcaption}

\DeclareCaptionLabelFormat{subcaptionlabel}{\normalfont(\textbf{#2}\normalfont)}
\newcolumntype{b}{X}
\newcolumntype{s}{>{\hsize=.5\hsize}X}

\begin{document}

\title{Leveraging Large Language Models for Integrated Satellite-Aerial-Terrestrial Networks: Recent Advances and Future Directions}
\author{Shumaila Javaid, \IEEEmembership{Member, IEEE}, Ruhul Amin Khalil, \IEEEmembership{Member, IEEE}, Nasir Saeed, \IEEEmembership{Senior Member, IEEE}, Bin He, \IEEEmembership{Senior Member, IEEE}, and Mohamed-Slim Alouini, \IEEEmembership{Fellow, IEEE}\\
\thanks{Shumaila Javaid and Bin He are with the Department of Control Science and
Engineering, College of Electronics and Information Engineering, Tongji
University, Shanghai 201804, China, and also with Frontiers Science Center
for Intelligent Autonomous Systems, Shanghai 201210, China. Email: shumaila@tongji.edu.cn; hebin@tongji.edu.cn.}
\thanks{Ruhul Amin Khalil and Nasir Saeed are with the Department of Electrical and Communication Engineering, College of Engineering, UAE University, Al-Ain 15551, UAE e-mail: ruhulamin@uaeu.ac.ae; mr.nasir.saeed@ieee.org.}
\thanks{Mohamed-Slim Alouini is with the Computer, Electrical and Mathematical Science and Engineering (CEMSE) Division, King Abdullah University Of Science And Technology 23955, Saudi Arabia.}}

\maketitle

\begin{abstract}
Integrated satellite, aerial, and terrestrial networks (ISATNs) represent a sophisticated convergence of diverse communication technologies to ensure seamless connectivity across different altitudes and platforms. This paper explores the transformative potential of integrating Large Language Models (LLMs) into ISATNs, leveraging advanced Artificial Intelligence (AI) and Machine Learning (ML) capabilities to enhance these networks. We outline the current architecture of ISATNs and highlight the significant role LLMs can play in optimizing data flow, signal processing, and network management to advance 5G/6G communication technologies through advanced predictive algorithms and real-time decision-making.
A comprehensive analysis of ISATN components is conducted, assessing how LLMs can effectively address traditional data transmission and processing bottlenecks. The paper delves into the network management challenges within ISATNs, emphasizing the necessity for sophisticated resource allocation strategies, traffic routing, and security management to ensure seamless connectivity and optimal performance under varying conditions.
Furthermore, we examine the technical challenges and limitations associated with integrating LLMs into ISATNs, such as data integration for LLM processing, scalability issues, latency in decision-making processes, and the design of robust, fault-tolerant systems. The study also identifies key future research directions for fully harnessing LLM capabilities in ISATNs, which is crucial for enhancing network reliability, optimizing performance, and achieving a truly interconnected and intelligent global network system.
\end{abstract}


\section{Introduction}
\label{sec:introduction}

Integrated Satellite Aerial Terrestrial Networks (ISATNs) represent a unified framework that enhances communication capabilities across various sectors by seamlessly integrating satellite, aerial (e.g., Unmanned Aerial Vehicles  (UAVs) or drones), High Altitude Platforms (HAPs), and terrestrial communication networks \cite{zhu2021integrated,zhang2020survey}. This integration significantly extends coverage, offering reliable communication services to hard-to-reach areas, including remote, rural, and maritime regions where traditional terrestrial networks are unavailable or economically unfeasible \cite{wang2019convergence,chao2019space}.

Satellite networks expand reach by bridging communication gaps in remote and rural areas, while aerial networks provide adaptable and swift coverage enhancements, particularly useful in rapidly changing or temporary scenarios. Terrestrial networks, meanwhile, deliver robust and high-speed communication solutions in densely populated areas, ensuring high-performance connectivity where infrastructure is well-established. This integrated approach allows ISATNs to deliver reliable, continuous communication services crucial for supporting emerging technologies that require widespread and uninterrupted connectivity, such as the Internet of Things (IoT), smart cities, and post-disaster communications. ISATNs leverage the strengths of each network type to overcome individual limitations, resulting in a robust, efficient, and comprehensive communication infrastructure \cite{yao2018space,pervez2021joint,alsharoa2020improvement}.

However, the broad benefits of ISATNs in unifying different communication modalities into a cohesive system come with significant challenges. Traditional network management approaches often fall short when applied to ISATNs due to their complexity and the unique demands of integrating disparate technologies. Unlike conventional networks, ISATNs must coordinate between satellites, aerial platforms, and terrestrial systems, each with distinct operational characteristics and requirements. For instance, satellites cover large areas but suffer from high latency and limited bandwidth compared to terrestrial networks, which offer low latency and high data throughput but limited geographic coverage. Aerial platforms, such as drones, add another layer of complexity with their mobility and lower altitude operations.

The variability in network conditions across these components further complicates network management. Environmental factors, such as atmospheric conditions, affect satellite signal quality, while physical obstacles can impact terrestrial network performance. This variability can lead to potential issues with latency, where data transmission delays can disrupt time-sensitive applications. Data throughput can also be inconsistent, with available bandwidth fluctuations affecting service quality. Moreover, maintaining signal integrity becomes challenging as signals traverse different mediums, each with its own set of interference and degradation issues. This results in one of the crucial challenges of interference management faced by ISATNs. Cross-layer signal interference among satellite, aerial, and terrestrial components significantly disrupts communication, causing propagation effects such as signal attenuation, path loss, and multipath fading
\cite{yan2020interference,liang2021realizing,peng2022integrating}.

Seamless switching between satellite, aerial, and terrestrial networks (e.g., when a device moves from a rural area with satellite coverage to an urban area with dense terrestrial infrastructure) is another critical challenge, requiring highly responsive and reliable handover mechanisms.
Besides, the open nature of wireless communication inherent to these networks increases their vulnerability to eavesdropping and jamming attacks, where unauthorized interception and disruption of signals can occur more easily, affecting integrity and confidentiality across all network segments \cite{li2020distributed,cai2024privacy,han2022challenges}. Ensuring data security and user privacy across these diverse integrated networks is paramount, as the integration significantly expands the potential attack surface for malicious activities \cite{ahmad2022security,fadlullah2021smart}. The complex nature of ISATNs requires sophisticated automation for network monitoring, maintenance, and resource allocation, as traditional approaches often rely on manual interventions, which are inadequate for the scale and complexity of ISATNs.
Additionally, traditional network systems must incorporate robust fault tolerance mechanisms to meet the high-reliability requirements of ISATNs, especially in critical applications such as emergency services or military operations. Optimization of latency and bandwidth usage is also mandatory, as the extensive coverage of satellite networks comes at the expense of high latency and bandwidth utilization. Moreover, ISATNs must adapt to dynamically changing network loads and expanding infrastructure while maintaining performance and reliability, posing significant logistical and technical challenges \cite{fang20215g,liu2018space}.

\begin{figure}[h!]
\begin{center}
\includegraphics[width=1\columnwidth]{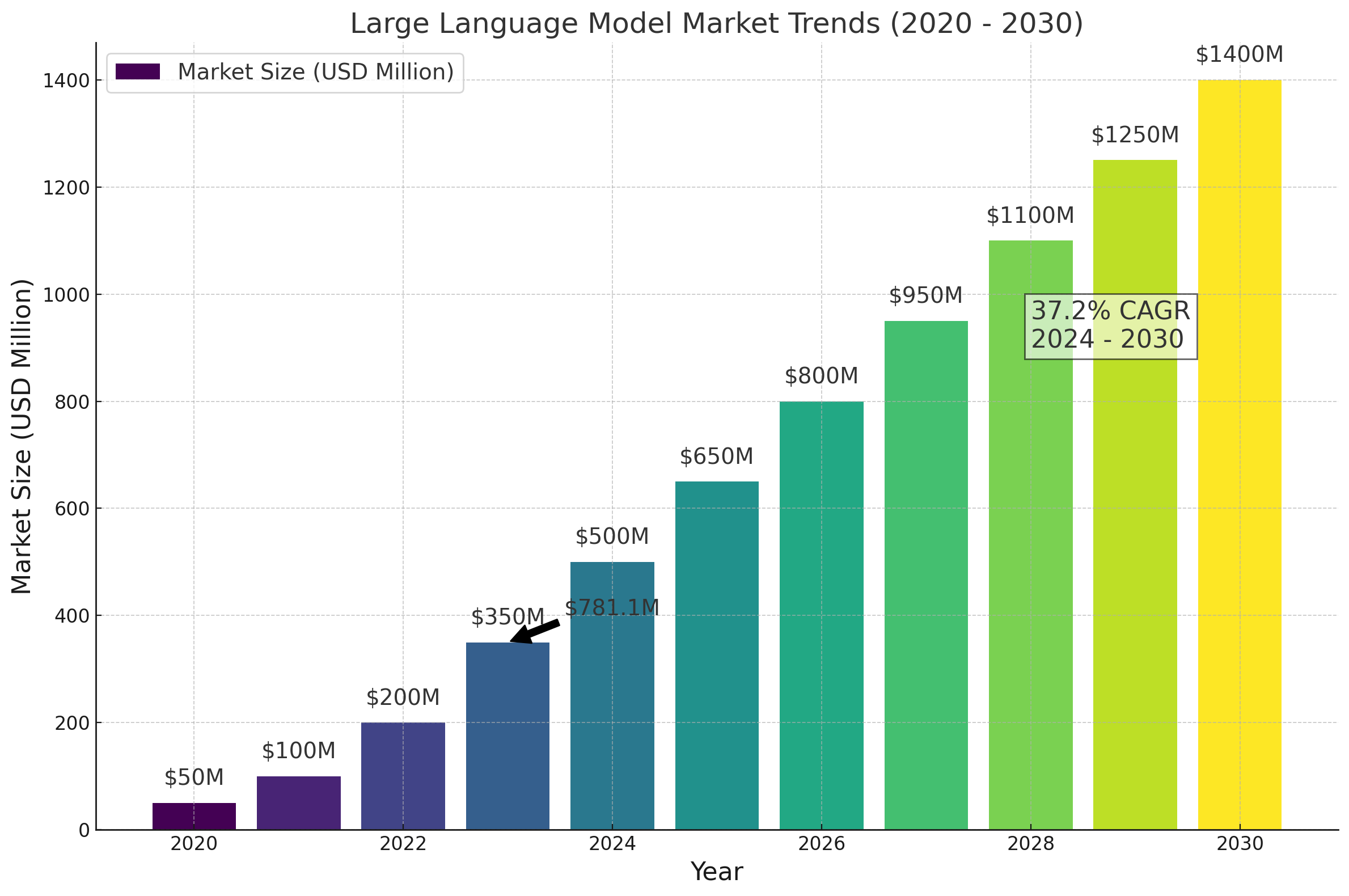}
\caption{Projected growth of the LLM market trends by 2030.}\label{LLMmarkettrends}
\end{center}
\end{figure}

Addressing the challenges and limitations of ISATNs requires innovative approaches in network architecture design, management, and security \cite{lee2022integrating}. Efficient spectrum management techniques among different layers can prevent overlap and ensure optimal usage \cite{liang2021realizing,peng2022integrating}. Incorporating CR technology can allow devices to automatically detect available channels in the wireless spectrum and adjust transmission or reception parameters to improve spectrum efficiency \cite{jia2016broadband,kolawole2017performance,zhu2019cooperative}. Advanced network management systems can enable heterogeneous network integration by incorporating various protocols, standards, and technologies. Moreover, ISATNs can integrate Software-Defined Networking (SDN) and Network Function Virtualization (NFV) to manage and reconfigure networks dynamically \cite{bi2019software,faheem2022performance,zhang2017software,yuan2023joint}. Network slicing can also support diverse needs, from high-speed internet browsing to high-priority emergency communications, by creating virtual networks on the same physical infrastructure.
Moreover, recent works have emphasized the role of Machine Learning (ML), Artificial Intelligence (AI), and edge computing in enhancing the efficiency and reliability of ISATNs through predictive analytics, network optimization, load offloading, and automated fault detection and recovery \cite{kato2019optimizing,al2023artificial,zhang2023ai}. Blockchain technology has garnered attention for its potential to strengthen security and enable decentralized network management by providing transparent and tamper-proof systems for handling complex network operations \cite{wang2021blockchain,han2022challenges,wang2021blockchain1}. Furthermore, Large Language Models (LLMs) have recently emerged as powerful tools for improving natural language processing tasks within these intelligent network systems. According to \cite{llmmarket2023trends}, there is significant growth in the U.S. LLMs market, projecting a rise from USD 50 million in 2020 to USD 1.4 billion by 2030, highlighting a robust Compound Annual Growth Rate of 37.2\% from 2024 to 2030 as shown in Fig. \ref{LLMmarkettrends}.

LLMs introduce a significant layer of innovation in network management, particularly within ISATNs. These advanced AI tools can process and generate human-like text based on extensive data analysis, enabling them to understand and predict complex network behaviors and requirements effectively. This capability is crucial for addressing the dynamic and multifaceted challenges inherent in ISATN environments \cite{chang2024survey,zhao2023survey}.
LLMs can transform the automation and optimization of network management tasks in ISATNs that traditional algorithms find too complex. For instance, LLMs can analyze historical network data to forecast traffic loads, identify potential faults, and recommend optimal resource allocations to maintain network reliability and efficiency in dynamic settings. Additionally, LLMs can aid in real-time decision-making by offering solutions or adjustments based on live data inputs from various network segments \cite{hadi2023large,hadi2023survey}.
Integrating LLMs into ISATN management can facilitate more effective communication between network layers and segments by processing natural language queries from network operators. Furthermore, LLMs can translate complex network metrics into actionable insights, enhancing operational transparency and control. The potential of LLMs to tackle ISATN management challenges could increase efficiency and significantly reduce the costs and workforce required for network supervision and troubleshooting \cite{min2023recent,hadi2023survey}.

Despite their vast potential, applying LLMs in ISATNs remains significantly underexplored. Therefore, this article delves into the role of LLMs in enhancing ISATNs. The primary motivation for this review is to address the pressing need for sophisticated solutions capable of managing the complexities associated with integrating diverse network components. This survey paper aims to provide a thorough understanding of how LLMs can mitigate existing challenges and maximize the capabilities of ISATNs by exploring the innovative ways in which LLMs can optimize data flow, enhance signal processing, and improve network management. This is essential for advancing global communication networks, ensuring robust and reliable connectivity, and facilitating the progression of next-generation communication technologies. As seamless, high-speed connectivity becomes increasingly crucial, comprehending and harnessing the potential of LLMs within ISATNs is imperative for substantially improving network efficiency and performance. This review addresses current knowledge gaps and establishes a foundation for future research, providing a vital resource for academics and industry professionals aiming to drive innovation in communication networks.
The following section reviews existing surveys to provide a foundation for understanding the current state of research in this field.

\newcolumntype{C}{>{\arraybackslash}X} 
\setlength{\extrarowheight}{1pt}
\begin{table*} [htp!]
 \caption{Summary of existing surveys related to LLMs in diverse domains.}
\label{table00}
 \begin{tabularx}
{\textwidth}{|s|s|b|b|b|}
\hline
\hline
\textbf{References} & \textbf{Research Focus} & \textbf{Findings and Technologies  Discussed} \\
\hline
\cite{wang2024survey} & Autonomous agents based on LLMs  & Detailed study on the design, implementation,  and assessment of agents using LLMs across various sectors    \\
\hline
\cite{xi2023rise}  & LLM-based AI agents development & Functions of LLMs in artificial general intelligence and development of a novel framework to improve the working of LLM in dynamic environments \\
\hline
\cite{wang2023aligning} & Adapting LLMs to meet human & Address issues such as misinterpretations, biased results, and errors in facts within content produced by LLM models. In addition, it offers a thorough examination of technologies that improve the alignment of LLMs with human expectations  \\
\hline
\cite{zhu2023survey} & Deployment challenges of LLMs & Investigates techniques for model optimization, including quantization, pruning, and distillation aimed at enhancing the efficiency of LLMs \\
\hline
\cite{gao2024llm} & Challenges of LLM & Examines the difficulties associated with LLMs,  focusing on dataset management, dependency on tokenizers, and the cost of pre-training \\
\hline
\cite{kaddour2023challenges} & LLM's limitations & Presents the operational difficulties faced by LLMs, including design shortcomings and inefficiencies   \\
\hline
\cite{sun2024generative} & generative AI in UAV systems & Examines generative AI technologies aimed at improving communication, networking, and security within UAVs  \\
\hline
\cite{liu2024generative} & generative AI in UAV swarms & Studies the challenges and potential of generative  AI in improving coordination and capabilities of UAV swarms \\
\hline
\cite{javaid2024large} & LLMs integration in UAV & Explores the integration of UAVs with LLMs to enhance their autonomous capabilities through  advanced AI and ML algorithms \\
\hline
This paper & LLM-based ISATNs & Explores the integration of LLMs into ISATNs, examining their impact on operational efficiency,  5G/6G advancements, network management strategies,  and addressing technical challenges. It also identifies future research directions for enhancing  communication reliability and network performance of LLMs-integrated ISATNs \\
\hline
\hline
\end{tabularx}
\end{table*}

\subsection{Related Surveys}
The growing potential of AI for enhancing network operation management, optimizing signals, automating fault detection, understanding and generating natural language, powering sophisticated chatbots, and improving machine translation systems has prompted the inclusion of several recent studies \cite{gao2023retrieval,mialon2023augmented,schwartz2023enhancing}. These studies focus on LLM architectures \cite{yin2023survey}, training processes, fine-tuning \cite{zhang2023instruction}, logical reasoning \cite{huang2022towards,liu2024large}, and other related challenges to address the limitations for broad adoption of LLM-based systems across all domains.
In \cite{wang2024survey}, the authors investigated LLM-based autonomous agents, focusing on their construction, application, and evaluation for autonomous decisions without human intervention. These agents interact with their environment and other users by integrating advanced AI techniques for enhanced communication, problem-solving abilities, and data analysis in various domains such as social science, natural science, and engineering. Similarly, \cite{xi2023rise} provided an in-depth review of LLM-based AI agents, exploring their role in advancing artificial general intelligence. The study concluded that LLMs are transformative for creating versatile AI agents due to their sophisticated language understanding and generation capabilities, significantly enhancing autonomous tasks in dynamic environments.
\cite{wang2023aligning} critically examined the challenges and advancements in aligning LLMs with human expectations to address issues such as misunderstanding instructions, factual inaccuracies, and biased outputs. The authors conducted an in-depth analysis of technologies (e.g., NLP benchmarks and human annotations) and training methodologies (e.g., supervised fine-tuning and human preference training) to enhance LLM understanding and response generation. In \cite{zhu2023survey}, the challenges of deploying LLMs in resource-constrained environments were explored, focusing on model compression techniques like quantization, pruning, and knowledge distillation to improve efficiency and practicality.
Further studies, such as \cite{gao2024llm,kaddour2023challenges}, examined the limitations of LLMs, including dataset management, tokenizers, high costs of pre-training and fine-tuning, inference time delays, limited text input lengths, and inadequate experimental designs. Additionally, \cite{sun2024generative} explored the role of generative AI in enhancing the communication, networking, and security features of UAV systems. They introduced a generative AI framework to improve UAV network performance. Similarly, \cite{liu2024generative} discussed the applications, challenges, and opportunities of generative AI in optimizing the coordination and functionality of UAV swarms, highlighting the significance of techniques such as generative adversarial networks, variational autoencoders and generative diffusion models in UAV operations.
In another study, \cite{javaid2024large}, the authors reviewed LLM architectures for UAV integration, highlighted state-of-the-art LLM-based UAV applications, and identified opportunities for enhancing data analysis, decision-making, and emergency response capabilities of LLM-based UAV systems.

The integration of LLMs with ISATNs appears underrepresented in the literature. This preliminary work focuses on integrating LLMs into ISATNs and discussing potential applications, opportunities, and future directions. It aims to lay the groundwork for future research and explore how LLMs can enhance communication and operational efficiencies within these complex network systems. This gap in the literature presents a unique opportunity to explore how advanced language processing capabilities can contribute to the evolution of ISATNs. Table \ref{table00} overviews existing LLM surveys, summarizing their key focus areas and discussed technologies.

\subsection{Contributions of the paper}
The contributions of this paper are summarized as follows:
\begin{itemize}
\item First, we examine the integration of LLMs within ISATNs, emphasizing the pivotal role of LLMs in boosting operational efficiency, advancing 5G/6G-based ISATNs, and overcoming traditional bottlenecks in data transmission and processing. This comprehensive study demonstrates how LLMs can optimize data flow, enhance signal processing, and implement sophisticated network management strategies, significantly transforming the capabilities and efficiency of ISATNs.
\item We then perform an in-depth exploration of complex network management challenges faced by ISATNs, highlighting the critical need for advanced strategies in resource allocation, traffic routing, and security management to ensure seamless connectivity and optimal performance under variable conditions. The study delves into the complexities of synchronizing diverse network components, managing real-time data-driven resource distribution, and dynamically adjusting traffic routes to mitigate bottlenecks while maintaining uniform security standards across various network environments.
\item Subsequently, we discuss how LLM integration into ISATNs can transform diverse applications. We identify and methodically address significant technical challenges associated with integrating LLMs into ISATNs, including issues related to data integration, scalability, latency in decision-making processes, and the need for robust, fault-tolerant designs. This contribution is crucial for laying the groundwork for future implementations and enhancements in network technologies.
\item Finally, we discuss future research directions aimed at harnessing the capabilities of LLMs within ISATNs. This includes developing specialized LLMs for real-time wireless channel management, advancing dynamic data integration techniques, and implementing adaptive strategies for network management. These initiatives aim to enhance communication reliability, optimize network performance, and ensure efficient, seamless integration across diverse network components.
\end{itemize}

\subsection{Organization}
\begin{figure*}[htp]
\begin{center}
\includegraphics[width=0.825\textwidth]{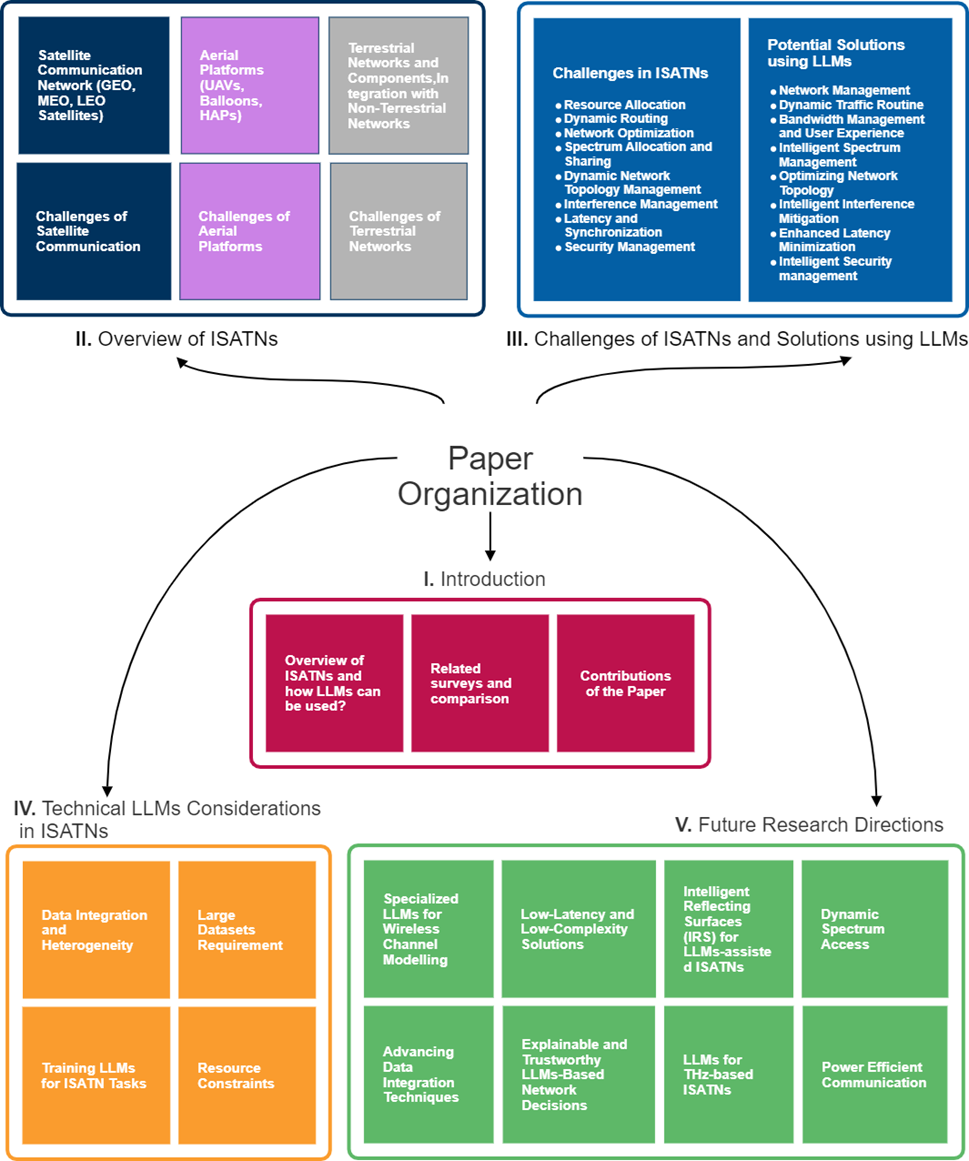}
\caption{Overall organization of this survey paper.} \label{organization}
\end{center}
\end{figure*}

The rest of the paper is organized as follows. Section \ref{sec:02} discusses the overview of ISATNs, detailing the components and functionalities of satellite networks, aerial platforms, and terrestrial networks and the unique challenges each faces. This includes an in-depth look at the integration of these networks and the potential improvements LLMs can bring. Section \ref{sec:03} provides the list of challenges of ISATNs, including resource allocation, traffic routing, and network optimizations, followed by possible solutions using LLMs. Section \ref{sec:04} discusses technical LLM considerations for network optimization, resource allocation, and management in ISATNs.  Section \ref{sec:05} explains the future research direction for effectively utilizing LLMs in ISATNs and identifies critical areas for future exploration and development. The concluding remarks are provided in Section \ref{sec:06}. Fig. \ref{organization} provides a detailed illustration of the various sections of this survey paper.

\section{Overview of ISATNs}\label{sec:02}
\begin{figure*}[htp]
\begin{center}
\includegraphics[width=1\textwidth]{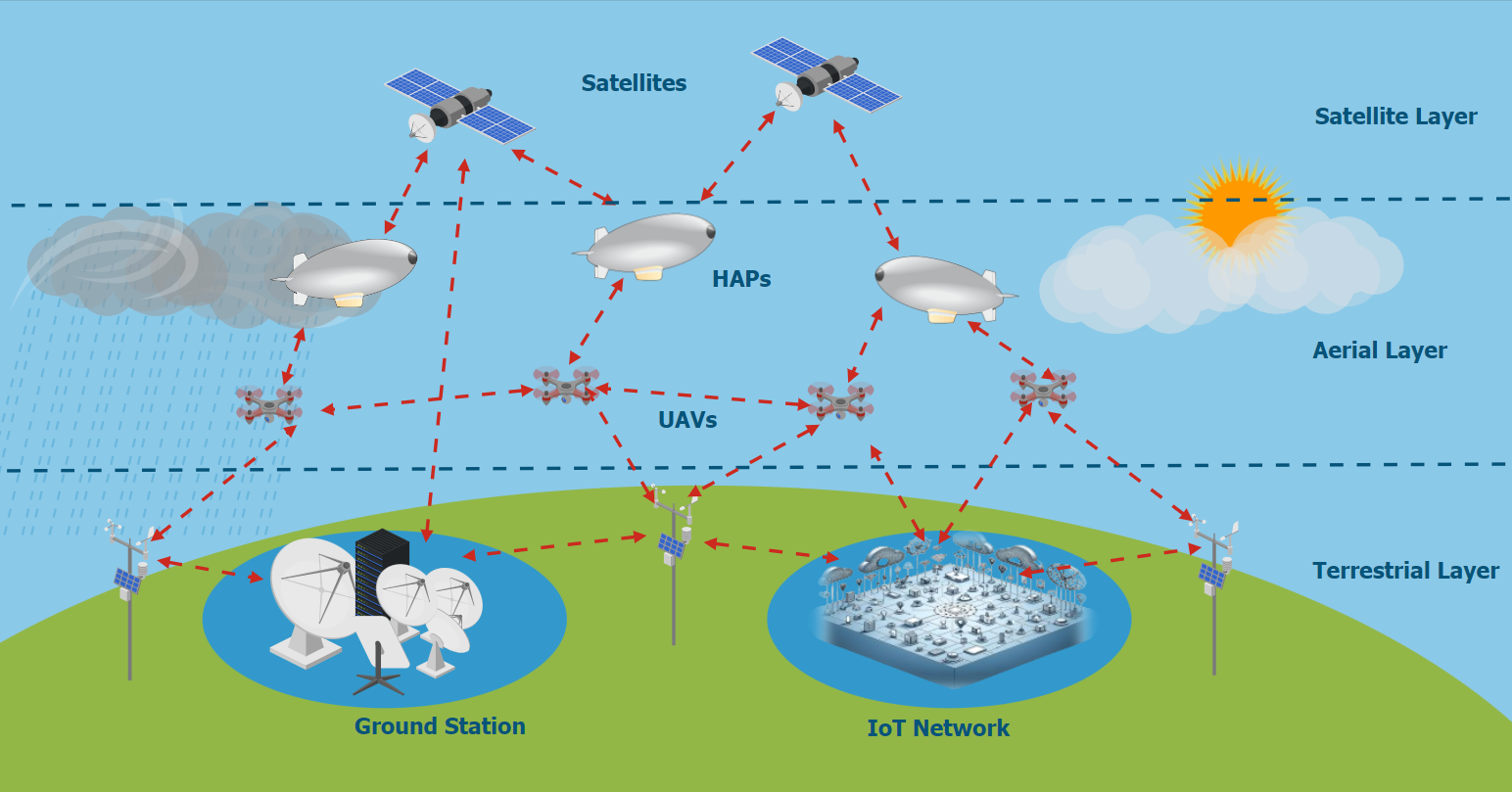}
\caption{A layer-by-layer architecture of ISATNs.} \label{fig:basicarc}
\end{center}
\end{figure*}

The transformative advancement in global communications led by ISATNs provides resilience, efficiency, and adaptive network infrastructure designed to meet current and future connectivity needs. This integration ensures that ISATNs can maintain connectivity during infrastructure disruptions such as natural disasters, power outages, and large-scale events, utilizing dynamic resource allocation to improve service quality and reduce operational costs. This section delves into the critical role of each component within ISATNs, highlighting their unique contributions and the potential for integrating LLMs to enhance their functionality. Fig. \ref{fig:basicarc} provides a layer-by-layer architecture of an ISATN.

\subsection{Satellite Communication Networks}
Satellites are pivotal for ISATNs to extend connectivity, covering broad geographical areas where terrestrial infrastructure is absent or too costly to establish. Their evolution from essential voice communication and broadcasting to complex roles like providing broadband Internet and global positioning services underscores the importance of selecting the appropriate satellite orbit \cite{kodheli2020satellite,maral2020satellite}. This selection is crucial to support diverse static and mobile platforms, including ships, airplanes, vehicles, emergency base stations, and inter-satellite links. The chosen orbit influences satellite service performance and suitability for various demands \cite{su2019broadband,al2022survey}. As such,
Geosynchronous Earth Orbit (GEO) satellites, positioned at 36,000 km, offer extensive coverage suitable for broadcasting and fixed satellite services. Their fixed positioning above a point on Earth makes them ideal for services requiring comprehensive, continuous coverage. However, their higher latency limits their use for broadband applications needing faster response times.
Medium Earth Orbit (MEO) satellites operate between 2,000 km and 36,000 km, with a balanced coverage area and latency. This makes them versatile for various services, such as newer broadband applications that require moderate latency. Low Earth Orbit (LEO) satellites, situated between 500 km and 2,000 km, play a crucial role in modern ISATNs due to their significantly lower latency and reduced signal attenuation. These attributes make LEO satellites ideal for delivering broadband Internet and MSS. The rapid development and strategic deployment of LEO satellites have transitioned satellite roles from traditional applications to providing robust, high-speed Internet services. Their proximity to Earth enables faster data services, which are essential for reliable Internet connectivity on mobile platforms and in remote locations. This adaptability meets the increasing demand for quick, dependable Internet access across diverse environments, highlighting the evolution of satellite networks as critical components of global communication infrastructure \cite{li2017service,furqan2022satellite}.

Integrating satellites in ISATNs significantly enhances network reliability by providing alternative communication pathways crucial during terrestrial network failures or overloads. This redundancy is vital for maintaining continuous service availability and supporting critical communication needs across healthcare, emergency services, and government operations. The diverse orbits of satellites (GEO, MEO, and LEO) allow for a layered redundancy approach tailored to specific coverage or latency requirements. Furthermore, satellites facilitate massive data transfers through broadcasting and multicasting capabilities, essential for media distribution, software updates, and emergency broadcasts over vast areas, especially where terrestrial infrastructure lacks reach or capacity \cite{ruan2020cooperative,sun2022integrated}.

For example, the O3b Network leverages MEO satellites to provide internet access to "the other 3 billion" in targeted latitudinal regions, demonstrating the strategic use of different satellite orbits to meet specific needs. LEO satellites, such as those from the Iridium network, offer global personal communication with advanced onboard processing and inter-satellite links. At the same time, Globalstar provides consistent access up to 70 degrees latitude—their rapidly expanding networks, like Starlink, plan to deploy nearly 42,000 LEO satellites to offer high-speed internet globally. Similarly, OneWeb and Telesat LEO enhance global coverage, aiming for commercial reach with their respective satellite constellations. This dynamic evolution underscores the crucial role of satellite communications in providing resilient connectivity for a wide range of applications, from rural broadband to international business, essential for the structure of ISATNs \cite{sanchez2017reliability,radhakrishnan2016survey}.

ISATNs significantly enhance 5G services by incorporating satellite communications, which address several limitations and challenges terrestrial and aerial-only networks face. Satellite communication extends 5G connectivity beyond the reach of terrestrial networks, ensuring that worldwide users can access high-speed, reliable mobile services, especially in regions where laying fiber optics or establishing ground infrastructure is impractical or cost-prohibitive \cite{zhang2020survey}. It also brings key enhancements to 5G capabilities, including Enhanced Mobile Broadband (eMBB) \cite{wan20184g,abdullah2021enhanced}, where satellites optimize operations by backhauling traffic from the network's edge and broadcasting content. They also provide direct connectivity in remote areas lacking terrestrial infrastructure, support Hybrid Multiplay, and facilitate mobile communications on moving platforms such as aircraft and vessels. For Massive Machine-Type Communication (mMTC), satellites enable IoT connectivity across extensive, often remote areas, supporting applications ranging from agriculture to fleet management and ensuring service continuity \cite{sharma2019toward,bockelmann2016massive}.

Moreover, regarding Ultra-Reliable and Low Latency Communications (uRLLC), satellites aid critical applications such as autonomous driving and remote surgery by offloading non-critical data to manage congestion and improve reliability \cite{ji2018ultra,lin2018ultra}. The strategic integration of satellite networks within ISATNs addresses comprehensive challenges related to coverage, reliability, and capacity, thus enhancing global access to 5G services and making ISATNs a comprehensive solution for modern telecommunications needs. By facilitating connectivity for IoT devices and supporting uRLLC, satellites help reduce latency and increase network reliability, ensuring a more uniform 5G experience across different regions and reducing the digital divide. Ultimately, the critical capabilities brought by integrating satellite communication into ISATNs extend, enrich, and ensure the resilience of the network infrastructure, making it indispensable for a truly integrated global communication system \cite{zhang2020survey,fang20215g}.

Despite advancements in 5G networks, including higher data rates, lower latency, and enhanced connectivity, several challenges persist. As the focus shifts to developing 6G within ISATNs, satellite communication will play a pivotal role in addressing these shortcomings and extending capabilities. The envisioned 6G network aims to achieve data rates from 100 Gb/s to 1 Tbps, reduce latency to 0.1 ms, and improve other metrics such as positioning accuracy, energy efficiency, and connectivity density. Integrating satellite networks with terrestrial frameworks is crucial to meet these ambitious targets and ensure global connectivity \cite{zhu2021integrated,sattarzadeh2021satellite}.

The future 6G architecture is expected to seamlessly incorporate both terrestrial and non-terrestrial networks, including satellites, enabling economically viable and ubiquitous network coverage. This will extend the reach of network services to all global users and enhance the network's capacity to support a broader range of applications and scenarios outlined for 6G, such as ubiquitous mobile ultra-broadband, ultra-high-speed-with-low-latency communications, and ultra-high data density. With their extensive coverage capabilities, Satellites are particularly seen as a solution to overcome the geographical and economic barriers of terrestrial network expansions. The envisaged satellite-terrestrial network for 6G will include a mix of GEO, MEO, and LEO satellites, providing comprehensive global coverage and supporting an integrated approach to global broadband access, essential for IoT and other advanced applications anticipated with 6G \cite{giordani2020non,azari2022evolution}.

\subsubsection{Challenges of Satellite Communication Networks within ISATNs}
In satellite communications within ISATNs, one of the primary challenges is propagation delay. GEO satellites exhibit significant delay due to their high orbital altitude, and even though LEO satellites have reduced delay compared to GEO and MEO, it still impacts applications requiring real-time responses, such as certain aspects of 5 G's uRLLC. Another crucial challenge is signal attenuation caused by atmospheric conditions (e.g., rain fade and ionospheric disturbances). This leads to intermittent connectivity and necessitates more robust and expensive ground equipment to maintain reliable links. In addition, bandwidth limitations due to spectrum allocation constraints present significant regulatory and technical challenges. Satellites have limited bandwidth, and efficiently allocating sufficient bandwidth to meet the high data demands of modern applications while avoiding interference with terrestrial networks is challenging \cite{wei2021hybrid,al2022next,yan2020interference}.

Launching and maintaining satellite networks also involve significant initial and operational costs. While costs have decreased over time, especially with the advent of smaller satellites and reusable launch vehicles, they still represent a considerable investment. Integrating satellite systems with terrestrial and aerial networks requires complex coordination to manage handoffs and ensure seamless service across different network layers and technologies, including synchronization, network architecture, and consistent service quality. Furthermore, satellite communications are vulnerable to security threats, such as signal jamming and eavesdropping. This makes security a critical issue, especially for sensitive applications in military, government, and critical infrastructure sectors \cite{tedeschi2022satellite}. Another challenge is the increasing risk of orbital debris, particularly with the growing number of LEO satellites, which raises concerns about space debris and potential collisions. Effective traffic management and debris mitigation strategies are crucial to safeguard satellite operations \cite{anttonen2021space}. Additionally, satellite operations must adhere to complex international regulations governing space utilization, including frequency use, orbital slots, and environmental impacts \cite{wu2019satellite,murtaza2020orbital}.

While 5G and the upcoming 6G also introduce numerous challenges, 5G has high latency limitations, especially with GEO satellites and complex integration with terrestrial networks are prevalent, alongside spectrum interference and bandwidth limitations that may not consistently support 5G's high-speed requirements. As the industry progresses towards 6G, these challenges intensify with the exploration of higher frequency bands, such as Terahertz (THz), which involves atmospheric absorption and requires advanced technology. Both 5G and 6G demand sophisticated network slicing, increased energy efficiency, and significant advancements in satellite technology and network architecture to achieve global coverage with reliable and low-latency communications. Addressing these challenges necessitates continuous technological innovation, alongside proactive regulatory and legal adjustments to ensure comprehensive and secure global connectivity to fully realize the potential of satellite communications within ISATNs to support the globally evolving demands of network services \cite{gustavsson2021implementation,saeed2021point,azari2022evolution}.

\subsection{Aerial Platforms}
Aerial platforms are vital components of ISATNs, significantly enhancing network flexibility and coverage. These platforms include airborne technologies such as UAVs, balloons, and other HAPs, offering several advantages over traditional satellite and terrestrial networks. They provide reduced latency and greater deployment flexibility, essential for real-time communication applications like live video streaming or emergency responses. Additionally, aerial platforms can be swiftly deployed and tailored to meet specific needs, offering versatile responses to temporary coverage requirements or sudden increases in network load, unlike satellites that require extensive planning and fixed orbits \cite{javaid2023communication,khan2022emerging,kurt2021vision}.
UAVs are particularly valuable for ISATNs due to their mobility and flexibility. In emergencies where terrestrial networks are damaged or overwhelmed, such as natural disasters, UAVs can be rapidly deployed to establish temporary communication networks, providing essential services to rescue teams and affected populations. UAVs can also enhance network capacity during large public events where existing networks may be insufficient, ensuring adequate coverage and maintaining service quality \cite{khalil2023uavs,mozaffari2019tutorial,mohsan2022towards}.
On the other hand, HAPs, operating in the stratosphere, offer extended coverage for longer periods compared to UAVs. They are ideal for prolonged situations like scientific expeditions in remote areas or extended disaster relief efforts. These balloons can also be equipped with sensors for weather forecasting, environmental monitoring, and other scientific purposes, benefiting from their extensive coverage area. Other HAPs, such as airships or pseudo-satellites, can remain aloft at altitudes lower than traditional satellites, providing semi-permanent coverage and maintaining stable, reliable communication links over specific areas for months. This stability and comprehensive visibility make them perfect for surveillance, border control, and maritime monitoring, providing essential real-time data and communication support \cite{zhou2020overview,mershad2021cloud,du2019station}.
Integrating aerial platforms into ISATNs offers rapid deployment, scalability, and cost-effectiveness, especially in harsh or remote environments. Their ability to carry diverse payloads for communication and data collection makes them highly effective in addressing specific, time-sensitive, and location-based challenges. This enhances service reliability and accessibility across varied terrains and circumstances while being more economical to deploy and maintain than traditional satellite networks.

These aerial benefits are particularly beneficial for extending 5G and 6G coverage into rural and remote areas, ensuring broader network accessibility, and eliminating coverage 'dead zones.' The quick deployment of these platforms in hotspots to temporarily boost network capacity during emergencies and significant events is essential for maintaining high data rate communications, a core feature of 5G and 6G networks. 5G and 6G are set to expand uRLLC, which is critical for autonomous driving, telemedicine, and industrial IoT applications. Aerial platforms can provide robust and rapid communication links where instantaneous data transfer is vital, especially in areas where ground infrastructure might not be reliable \cite{geraci2022will,mozaffari2021toward}.

Moreover, the aerial platforms can also serve as moving nodes that provide backhaul connectivity from ground-based networks to satellite systems. This is especially useful in dense urban areas or regions where laying physical infrastructure is challenging or expensive. By facilitating efficient backhaul, aerial platforms help manage the increased data loads expected with 5G and 6G technologies. With the shift towards edge computing in 5G and 6G to reduce latency and network congestion, aerial platforms can be equipped with computing resources to process data closer to end-users. This capability is beneficial for real-time analytics, AI-driven services, and localized data processing, reducing the burden on core networks \cite{tezergil2022wireless,zhang2020beyond}.

Given their flexibility and ease of deployment, aerial platforms are ideal for testing and developing new 6G technologies in a controlled yet realistic environment. They can simulate different network scenarios and conditions, aiding in the research and development of 6G technologies before full-scale terrestrial deployment. Moreover, by integrating these aerial platforms into ISTANs, network operators can leverage their unique capabilities to address the specific demands of 5G and 6G networks, ensuring broader and more reliable coverage and supporting the advanced features these next-generation networks aim to offer.

\subsubsection{Challenges of Aerial platforms within ISATNs}
Aerial platforms within ISTANs face several challenges that must be addressed for optimal integration and operation. One of the significant challenges is the technical limitations of aerial platforms due to their limited battery life, payload capacity, and need for durable design to withstand various environmental conditions that impact their operational efficiency. The operation of aerial platforms, especially UAVs and balloons, highly depends on weather conditions, and adverse weather (e.g., high winds, rain, or extreme temperatures) substantially hinders their deployment, performance, and safety, limiting their availability and reliability \cite{javaid2023communication,mozaffari2019tutorial}. Balloons or HAPs also face hurdles in maintaining stable flight patterns and handling harsh atmospheric conditions. Communication Interference is another critical challenge as aerial platforms rely heavily on wireless communication that is susceptible to interference from various sources, including terrestrial networks and other aerial devices, and managing this interference to ensure reliable and secure communication is a significant technical hurdle \cite{kurt2021vision,fu2021collaborative}.

Moreover, integrating aerial platforms into existing ISTAN architectures poses scalability challenges. These include technical integration into network infrastructure, consistent and seamless communication across satellite, aerial, and terrestrial components, and expanding operations to cover extensive geographical areas without degrading service quality. In addition, aerial platforms, particularly drones and HAPs, operate in airspace regulated by national and international authorities. Navigating these regulations, which can vary significantly between regions and countries, presents a complex challenge, including issues related to flight permissions, safety standards, and privacy concerns. Aerial platforms are also vulnerable to security threats; thus, protecting the data transmitted and the platforms themselves is crucial, requiring robust security protocols and constant monitoring \cite{fotouhi2019survey,stocker2017review}.

Furthermore, since aerial platforms are considered cost-effective in specific applications, deploying and maintaining a fleet of UAVs, balloons, or other HAPs can be expensive due to the initial technological outlay and ongoing expenses related to operations, maintenance, and regulatory compliance. The environmental impact of deploying aerial platforms, particularly regarding emissions and wildlife disruption, needs consideration. Ensuring these platforms are environmentally sustainable is increasingly essential as regulatory and societal expectations around environmental protection grow \cite{jang2020cost,hu2020risk}.

In the context of aerial platforms for deploying and enhancing 5G and emerging 6G networks to deliver uRLLC, eMBB, and mMTC presents specific challenges unique to these next-generation networks. For instance, achieving the high data rates and extensive network coverage promised by 5G and 6G involves complex coordination between ground-based infrastructure and aerial platforms, such as drones and HAPs. The dynamic nature of aerial platforms requires adaptive beamforming technologies and highly responsive network management systems to ensure stable and efficient communication links. In addition, the density of aerial platforms needed to meet network demands can lead to increased risk of collisions and interference, demanding sophisticated traffic management solutions. As these networks also drive towards higher frequency bands, such as mmWave and sub-THz, aerial platforms must contend with higher propagation losses and more significant attenuation effects, which could limit their effective range and reliability. Addressing these technical complexities is critical to fully harness the potential of aerial platforms in enhancing 5G and 6G network capabilities within ISTAN frameworks \cite{duong2022uav,renga2022can,palanci2022high}.

\subsection{Terrestrial Networks}
Terrestrial networks form the backbone of modern communication systems, consisting primarily of ground-based infrastructure. These networks are crucial in providing most connectivity services to urban and suburban populations. Unlike satellite and aerial networks, which are typically used to supplement or extend coverage, terrestrial networks deliver the primary connectivity for most day-to-day communications and internet services \cite{wang2019convergence,rinaldi2020non}. The components of terrestrial networks include cellular networks, fiber optic cables, and other ground-based communications infrastructure such as fixed wireless access, digital subscriber line, and cable systems \cite{skubic2017optical,willner2019optical,kuang2018terrestrial}.

Terrestrial networks are especially vital in densely populated urban and suburban areas with the most significant demand for high-speed, high-capacity connectivity. These networks, particularly those utilizing 5G technologies, are designed to handle many simultaneous connections per square kilometer. This capacity is crucial in urban areas where thousands of devices, from smartphones to IoT devices, require constant connectivity. For applications requiring real-time response, such as autonomous driving, virtual reality, and industrial automation, the low latency offered by terrestrial networks is indispensable. Moreover, ground-based infrastructure is generally more robust against weather-related disruptions than satellite or aerial platforms, making this reliability essential for emergency services, healthcare, and other critical applications \cite{vaezi2022cellular,liu2021robotic}.

The terrestrial network forms the foundational infrastructure in ISTANs for integration with non-terrestrial networks, such as satellites and UAVs, for offering 5G and 6G services. This integration aims to enhance coverage, connectivity, and performance across environments and use cases, including IoT and mobile edge computing. The development and evolution of these networks are characterized by the use of advanced technologies such as Millimetre Wave (mmWave) frequencies, ML for network management, and network virtualization. Integrating terrestrial and non-terrestrial networks in enhancing 5G and 6G is critical in forming a comprehensive, seamless communication infrastructure. The mmWave frequencies, ranging from 30 to 300 GHz, are instrumental in achieving the high data rates essential for next-generation applications. The integration of mmWave technology allows for the exploitation of wide bandwidths, thereby significantly increasing the capacity and reducing latency in urban and densely populated areas. However, the high frequency also means mmWave signals have a shorter range and are more susceptible to physical obstructions, which is where non-terrestrial networks (e.g., satellites, UAVs, and HAPs) play a role by enhancing coverage and reliability, especially in rural or challenging terrains \cite{fu2021collaborative,gao2024cooperative}.

On the other hand, network virtualization techniques abstract hardware resources into a software-based virtual network. This transformation allows for more dynamic and flexible network management, enabling operators to deploy and scale services more rapidly and efficiently. Network virtualization in an integrated terrestrial and non-terrestrial network setup ensures that resources are used optimally, with traffic loads balanced and managed across various segments (terrestrial, aerial, and satellite) depending on demand and network conditions. Seamless integration of terrestrial and non-terrestrial networks requires sophisticated network management tools and protocols to ensure smooth operation and service continuity; thus, SDN and NFV are critical in this respect, providing centralized control and orchestration of network resources, which are especially beneficial in environments where network conditions can be highly variable and unpredictable. Through these technologies and integration strategies, the merged network can deliver more reliable services and a better user experience, supporting a wide range of modern applications and increasing overall network resilience against various connectivity and service delivery challenges \cite{bi2019software,yao2019si,lin2019virtualized}.

\subsubsection{Challenges of Terrestrial Networks within ISATNs}
Integrating terrestrial networks within ISTANs involves navigating several complex challenges. One of the primary challenges is managing interference between different network components. Terrestrial networks, especially those using high-frequency bands such as mmWave, face significant interference from satellite signals and vice versa. This interference can degrade the quality of service, leading to dropped connections and slowed data speeds, particularly in urban areas where the density of signals is higher. Achieving synchronization across terrestrial and non-terrestrial networks is another critical challenge. These networks must operate in harmony to provide seamless service delivery. This involves precise timing to ensure that data packets transmitted via different routes (satellite and terrestrial) arrive at their destination in the correct order and without excessive delay, which is vital for real-time applications such as Voice Over IP and streaming services \cite{chen2020system,boero2018satellite}.

The integration of terrestrial and non-terrestrial networks also faces regulatory hurdles. Different countries have varying spectrum-use regulations, and coordinating these can be challenging. Additionally, the competition for limited spectral resources can create conflicts between service providers, necessitating careful planning and negotiations. Moreover, integrating diverse network systems with their own protocols and standards adds substantial complexity, requiring advanced engineering solutions and often personalized adaptations to ensure compatibility across satellite, UAV, and terrestrial technologies. Cost is another crucial concern as installing, maintaining, and upgrading terrestrial network components and installing new ground stations requires significant investment. The need for ongoing maintenance and technological updates further accelerates this challenge \cite{pervez2021joint,xie2020satellite}.

Furthermore, with the broader network architecture, security becomes increasingly challenging. The integrated network's expanded attack surface requires robust security strategies to protect against various cyber threats (e.g., jamming attacks,
denial of service attacks, spoofing, eavesdropping, and Ransomware), ensuring user data's integrity and confidentiality across terrestrial and non-terrestrial components. These challenges demand innovative approaches and collaboration across industry stakeholders, regulatory bodies, and technology providers to ensure that ISTANs can deliver the envisioned benefits of enhanced connectivity and performance in the 5G and 6G era \cite{guo2021survey,yin2021physical,zhu2021integrated}. Table \ref{tab:istans} further highlights the critical components of ISATNs, providing a detailed description of each element, associated challenges, and the benefits of integrating LLMs to enhance their performance and reliability.

\newcolumntype{C}{>{\arraybackslash}X} 
\setlength{\extrarowheight}{1pt}
\begin{table*} [htp!]
 \caption{Overview of ISATNs entities, their description, challenges, and benefits of LLM integration.}
\label{tab:istans}
 \begin{tabularx}
{\textwidth}{|b|b|b|b|}
\hline
\hline
\textbf{ISATNs Components} & \textbf{Description} & \textbf{Challenges} & \textbf{LLM Integration Benefits} \\
\hline
\textbf{Satellite Networks} & Enhance network reliability by providing alternative communication pathways, crucial during terrestrial network failures or overloads. They facilitate massive data transfers through broadcasting and multicasting, essential for wide-area communication  & Propagation delay, signal attenuation, bandwidth limitations, high costs, complex integration, security  threats, risk of orbital debris, regulatory compliance & LLMs can optimize data flow, enhance signal processing, improve predictive maintenance, and manage network traffic effectively    \\
\hline
\textbf{Aerial Platforms}  & Enhances network flexibility and coverage. They offer reduced latency and greater deployment flexibility, which is crucial for real-time communication applications. UAVs provide rapid deployment in emergencies, while HAPs offer extended coverage for prolonged periods & Technical limitations such as limited battery life and payload capacity, dependence on weather conditions, communication interference, scalability challenges, regulatory issues, security threats, high operational costs, and environmental impact & Improve operational efficiency, enhance real-time decision-making, optimize data flow, manage network traffic effectively, predict maintenance needs, and enhance overall network reliability  \\
\hline
\textbf{Terrestrial Networks} & Terrestrial networks form the backbone of modern communication systems, providing the bulk of connectivity services in urban and suburban areas. Essential for high-speed, high-capacity connectivity, particularly in densely populated areas, and crucial for applications requiring real-time response such as autonomous driving and industrial automation & Managing interference between different network components; synchronization across terrestrial and non-terrestrial networks; regulatory hurdles; competition for limited spectral resources; cost of installation, maintenance, and upgrades; expanded attack surface and cybersecurity threats & Enhances real-time network management, improves traffic optimization, predicts maintenance needs, enhances data flow management, and increases overall network reliability and performance  \\
\hline
\hline
\end{tabularx}
\end{table*}

\section{Challenges of ISATNs and Solutions using LLMs}\label{sec:03}
The complexity of coordinating between terrestrial and non-terrestrial components requires advanced composition and synchronization to ensure seamless connectivity and service delivery.  The variability in network conditions across these diverse components complicates resource allocation and traffic management, posing difficulties in maintaining optimal performance. Furthermore, implementing consistent security protocols across varied network environments is challenging due to differing operational and regulatory standards. Addressing these complexities is essential for effective network management, as it directly impacts the network's reliability, efficiency, and security. This section highlights the major challenges in network management for ISATNs and explores how LLMs can provide innovative solutions. By leveraging LLMs, it is possible to optimize data flow, improve signal processing, and manage network traffic more effectively, thereby addressing the critical issues that impede seamless integration and performance in ISATNs.

\subsection{Challenges of ISATNs}
\subsubsection{Resource Allocation}
Resource allocation in ISATNs involves strategically distributing network capabilities across various terrestrial and non-terrestrial components. It is a complex task due to the diverse characteristics of these components. The allocation process must be highly adaptive due to the dynamic nature of network conditions influenced by satellite orbits, atmospheric conditions, and terrestrial network congestion. Resource allocation algorithms must continuously assess and adjust resource distribution based on real-time data. For example, satellite links have bandwidth capacities and latencies different from terrestrial links. Allocating resources efficiently requires adaptive strategies that respond to changing network conditions in real-time \cite{chan2022intelligent,zhu2018energy,chen2019efficient}.

Another primary challenge is to maximize network performance while minimizing costs and maintaining service quality across different regions and conditions. Resource allocation must balance the demand from users and applications with the available network resources, which can fluctuate due to the mobile nature of non-terrestrial components. Moreover, effective resource allocation also depends on integrating technologies such as 5G/6G, mmWave, and network virtualization, which help manage and allocate resources dynamically across the network. Addressing these challenges requires sophisticated management tools and algorithms that can make real-time decisions about resource distribution, ensuring optimal performance of the network as a whole. Moreover, prioritizing traffic and resources based on user demands and application requirements is essential. Policies must be developed to dictate allocated resources, ensuring critical services maintain high quality even under strained conditions \cite{fu2020integrated,peng2021hybrid,zhang2019joint}.

The current literature has emphasized integrating SDN and NFV in resource management to facilitate the virtualization of network functions and enhance resource allocation flexibility. However, achieving effective resource allocation in ISATNs requires technical solutions, strategic planning, and policy-making. This holistic approach is crucial to adapt continuously to the evolving demands of the network and the rapid advancement of technologies \cite{shi2019cross,zhang2017software,yuan2023joint,lin2019virtualized,lyu2020virtualized}.

\subsubsection{Dynamic Traffic Routing}
Dynamic traffic routing in ISATNs is crucial in efficiently managing network resources using advanced analytic tools that monitor real-time traffic data across the network to adopt fluctuating traffic patterns. Ensuring continuous service and optimized network performance when terrestrial paths become overloaded requires adaptive routing protocols to reroute data through alternative satellite or aerial links \cite{xu2022link,torkzaban2020join}. Dynamic routing is also essential for enhancing robustness by load balancing to prevent links or nodes from bottlenecking, which is particularly vital in ISATNs due to the variability in performance influenced by natural elements such as atmospheric conditions on satellite transmissions or urban infrastructures on terrestrial signals \cite{yin2023joint,lv2020routing}.

Current literature has focused on integrating ML and predictive modeling for adaptive routing to proactively adjust resources and routes in the network, smoothing out potential spikes in demand. The AI-advanced routing capability provides multiple data pathways to safeguard against failures. It prioritizes traffic based on the critical nature of applications, ensuring that high-priority services such as emergency communications and real-time gaming maintain superior quality of service. Intelligent routing decisions also consider other factors, including data type, service priority, and current network conditions, to optimize traffic flow \cite{tang2021deep,kato2019optimizing,nie2024routing,zhao2022orbital}.

State-of-the-art literature has also considered integrating coordination mechanisms that address the inherent differences in latency and bandwidth capabilities of ISTAN to ensure seamless service delivery between different components of ISTANs \cite{tang2020dynamically,li2020maritime}. However, with the evolution of ISATNs, the need for sophisticated traffic management systems is increasing to leverage cutting-edge technologies to anticipate and mitigate potential network issues before they impact performance. Scalability and flexibility are additional challenges for dynamic routing systems, which must accommodate increasing demands without compromising performance. Automated self-healing capabilities are needed to detect and rectify network failures autonomously, enhancing overall resilience. Energy efficiency has been an ongoing focus for adaptive routing algorithms for energy-efficient routing, optimizing routes to minimize energy consumption, which is crucial for sustainability in large-scale networks \cite{ruan2017energy,ruan2017energy}. Furthermore, the adaptive routing algorithms must comply with international standards for data routing to ensure robust security measures to protect data integrity and privacy for effective network management in ISATNs.

\subsubsection{Network Optimization}
Network optimization in ISATNs focuses on enhancing both network efficiency and performance to satisfy current communication demands. It involves several layers of strategy and technology aimed at fine-tuning the network to operate seamlessly across various environments. Current literature has put significant effort into developing sophisticated adaptive algorithms that dynamically manage and optimize network resources based on real-time data analysis. These algorithms are crucial for adapting to the fluctuating demands and diverse conditions characteristic of ISATNs. Although progress has been notable, the complexity and variability of ISATNs demand continuous focus and enhancement to ensure seamless integration of diverse technologies and extensive geographical coverage \cite{pervez2021joint,alsharoa2020improvement,khalil2020network,pan2020performance}.

In addition to adaptive algorithms, network optimization requires advanced predictive analytics to forecast potential network congestion and performance bottlenecks by analyzing trends and historical data. By anticipating issues before they occur, network administrators can proactively adjust resources, reroute traffic, or scale network capabilities, thus preventing disruptions and maintaining service quality \cite{abderrahim2022proactive,cheng20226g}. Load balancing techniques are also essential for the network optimization of ISATNs to evenly distribute traffic across all network resources, given that different components of ISATNs have varying capabilities and bandwidth availabilities. Thus, effective load balancing is also an ongoing challenge to ensure that no single part of the network bears too much load, which can lead to degradation in service and potential outages \cite{shahid2020load,tao2022traffic,li2020energy}. Furthermore, network optimization also needs to ensure robust security to safeguard the users from vulnerability and protect data integrity and user privacy across the network \cite{dai2020dynamic,ahmad2022security}.

\subsubsection{Spectrum Allocation and Sharing}
Spectrum allocation and sharing in ISATN are critical for optimizing available frequency bands and ensuring efficient communication across diverse network components. ISATN integrates satellite, aerial, and terrestrial systems with unique spectrum requirements and operational characteristics \cite{liang2021realizing, jia2020intelligent, zhu2017resource}. Satellite systems typically operate in higher frequency bands, such as C-band, Ku-band, and Ka-band, while terrestrial systems utilize lower frequency bands like sub-6 GHz and millimeter-wave bands. Aerial systems like drones and UAVs can operate in various frequency bands depending on their mission and operational requirements. This heterogeneity in spectrum usage among ISATN components poses significant challenges regarding interference management, resource allocation, and spectrum sharing \cite{li2023dynamic}. Traditional spectrum management techniques, which often rely on static allocation and fixed policies, are inadequate for addressing the dynamic and complex nature of ISATN. The constantly changing network topology, user mobility, and varying traffic demands necessitate more advanced and adaptive approaches to spectrum management \cite{liu2021spectrum}.

A key challenge in ISATN is the coexistence of multiple communication systems within overlapping frequency bands, which can lead to interference and reduced network performance. Interference can occur between different network components (e.g., satellite-to-terrestrial, aerial-to-satellite) or within the same component (e.g., inter-satellite interference, intra-terrestrial interference) \cite{biyoghe2021comprehensive}. This interference can degrade signal quality, increase bit error rates, and ultimately impact overall network capacity and user experience. Techniques such as Non-Orthogonal Frequency Division Multiplexing (NOMA) and Cognitive Radio (CR) have been proposed to improve spectrum-sharing efficiency in ISATN \cite{zhao2022integrated, han2023rate}. NOMA allows multiple users to transmit on the same carrier by distinguishing them based on power levels, enabling more efficient use of the available spectrum. However, NOMA requires careful power allocation and interference management to ensure reliable communication. On the other hand, CR enables dynamic spectrum access by sensing and adapting to the spectrum environment, allowing secondary users to opportunistically access underutilized frequency bands without causing harmful interference to primary users \cite{feng2017coordinated}. Nonetheless, CR implementation in ISATN faces challenges such as accurate spectrum sensing, efficient spectrum handoff, and coordination among heterogeneous network components.

\subsubsection{Dynamic Network Topology Management}
Another major issue in ISATNs is the dynamic network topology, which is essential for maintaining efficient and reliable communication across diverse network components \cite{lyu2024dynamic, han2022time}. The static approaches used in traditional topology management techniques cannot adapt to the rapidly changing conditions in ISATNs, leading to sub-optimal performance, resource wastage, or even service disruptions. For instance, a static topology configuration optimized for a particular traffic pattern may become inefficient when user mobility or traffic loads change, resulting in increased latency, congestion, or connectivity issues \cite{bi2019software}. Instead, advanced methods capable of dynamically adjusting the network topology in real time are necessary to optimize performance, reduce latency, and enhance overall network resilience. By continuously monitoring and analyzing various network parameters, such as link quality, resource utilization, and traffic patterns, these dynamic topology management techniques can identify potential bottlenecks or inefficiencies \cite{cao2022dynamic}. Based on this analysis, the network topology can be reconfigured by adjusting parameters such as routing paths, resource allocation, and network component configurations (e.g., satellite beam patterns and aerial vehicle positioning), improving network performance and resilience.

The dynamic nature of ISATN, which encompasses satellite systems in various orbits, aerial vehicles with varying mobility patterns, and terrestrial networks with fixed and mobile components, requires continuous adaptation to changing conditions \cite{dai2020dynamic}. These conditions include user mobility, varying traffic loads, and environmental factors such as weather conditions, terrain, and obstacles \cite{li2024dynamic}. Traditional topology management techniques that rely on static configurations or predefined rules are inadequate for handling these dynamic requirements, as they fail to capture the complex inter-dependencies and trade-offs between different network components.

\subsubsection{Interference Management}
Interference management in ISATN is a crucial challenge for ensuring seamless and reliable communication across diverse network components \cite{niu2024interference}. The integration of satellite, aerial, and terrestrial systems introduces various types of interference, including Self-Interference (SI), Mutual Interference (MI), clutter, and Cross-Link Interference (CLI). Effective management of these interferences is essential to maintain optimal network performance. Traditional interference management techniques in ISATN involve interference suppression, avoidance, and exploitation \cite{sharma2021performance}. For instance, beamforming and interference cancellation are employed to suppress unwanted signals. Meanwhile, dynamic spectrum allocation and multiple access techniques such as time division multiple access, frequency division multiple access, and NOMA help avoid interference by allocating distinct resources to different users.

Additionally, advanced methods like rate-splitting multiple access exploit interference to enhance spectral efficiency by treating some interference as noise and partially decoding it \cite{liu2023distributed}. However, these traditional techniques may not suffice to address the dynamic and complex nature of ISATN, where network topology, user mobility, and traffic patterns are constantly evolving. Effective interference management in ISATN requires advanced techniques that can adapt to these dynamic conditions and provide real-time solutions \cite{mao2022rate, coleman2023context}. Moreover, the coexistence of multiple communication systems within overlapping frequency bands and the diverse operational characteristics of satellite, aerial, and terrestrial components further exacerbate the interference challenges in ISATN. Interference can degrade signal quality, increase bit error rates, and ultimately impact overall network capacity and user experience, making it a critical issue that needs to be addressed effectively.

\subsubsection{Latency and Synchronization}
Latency and synchronization are critical factors influencing the performance of the ISATN \cite{dong2023delay}. Latency, the delay between sending and receiving data, can significantly impact user experience and network efficiency. In ISATN, latency issues are exacerbated by the diverse nature of the network components and the vast distances involved, particularly with satellite communications \cite{jiang2023network}. Satellite links typically experience higher latency due to long propagation distances, while terrestrial and aerial links exhibit relatively lower latency. This heterogeneity in latency characteristics across different network segments presents significant challenges in ensuring seamless communication and maintaining the Quality of Service (QoS) requirements for delay-sensitive applications \cite{wang2023time}.

Synchronization in ISATN is equally essential, as it ensures that data packets are transmitted, and received coordinated across different network segments. The heterogeneous nature of ISATN further complicates the synchronization process \cite{dandoush2024large, coleman2023context, feng2017coordinated}. Each network component may have different clock sources, timing references, and synchronization protocols, leading to potential timing misalignment and synchronization errors. These synchronization issues can result in data loss, packet reordering, and degraded network performance, particularly in applications requiring high precision and low latency, such as real-time video streaming and critical infrastructure monitoring \cite{proxet2023llm}. Effective synchronization mechanisms are crucial for maintaining the integrity and efficiency of data communication in ISATN, ensuring that data packets are delivered in the correct order and with minimal delays.

\subsubsection{Security Management}
Security management in ISATNs is paramount for protecting the network infrastructure and the data transmitted from potential threats. ISATNs require advanced encryption techniques to ensure that information remains confidential as it travels between network segments, including terrestrial, satellite, and aerial components. The encryption protocols for ISATNs must be continuously updated to counteract emerging cryptographic threats and vulnerabilities \cite{tedeschi2022satellite,wu2023threat}. State-of-the-art literature has introduced Intrusion Detection Systems (IDS) and Intrusion Prevention Systems (IPS) to monitor network traffic for suspicious activities for ISATNs. These systems are designed to detect potential threats in real-time and can automatically take actions to mitigate risks, such as blocking traffic from a malicious source or alerting network administrators to potential breaches \cite{li2020distributed,shrestha2021machine,guo2021survey}.

Existing studies \cite{xu2023anomaly,ahmad2022security} have also focused on regularly assessing network vulnerabilities. Through penetration testing and vulnerability assessments, security teams identify and rectify security weaknesses that attackers could exploit. This proactive approach helps fortify the network against known threats and prepares the network to withstand new types of attacks. Although significant advancements have been made in the security management of ISATNs, addressing the challenge of entirely securing these complex networks remains critical for ongoing research and development. As ISATNs expand and integrate diverse technologies across vast geographical areas, robust, up-to-date security measures to protect against evolving threats become increasingly essential. Table \ref{table22} briefly discusses various network management challenges and presents potential LLMs-based solutions.

\subsection{Potentials Solutions using LLMs}
\begin{figure*}[h!]
\begin{center}
\includegraphics[width=0.85\textwidth]{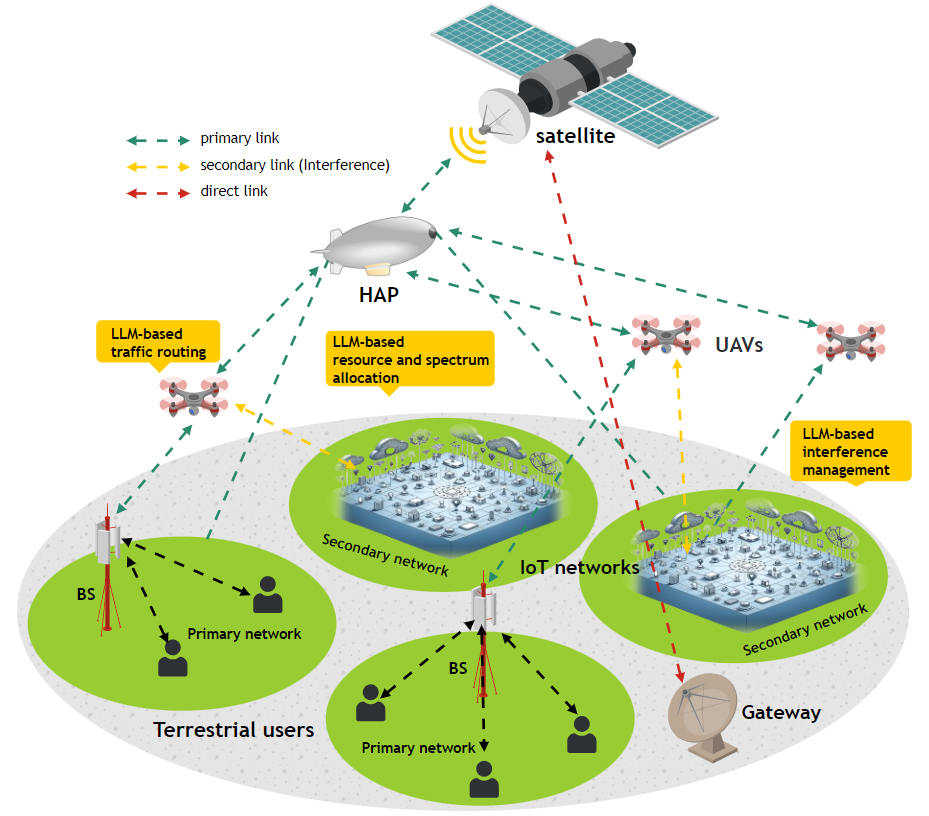}
\caption{Illustration of LLM-based solutions for various challenges in ISATNs.} \label{ISATNsChallengesLLM2}
\end{center}
\end{figure*}

In the previous subsection, we have introduced the potential challenges ISATNs face. In the following, we discuss the possible solutions for those challenges using LLMs. Fig. \ref{ISATNsChallengesLLM2} illustrates the LLM-assisted solutions for various challenges in ISATNs

\subsubsection{Network Management and Operations} LLMs can significantly enhance the efficiency and effectiveness of resource allocation in ISATNs through advanced data analysis and predictive capabilities. For instance, LLMs can identify patterns and trends in ISATNs that are not immediately apparent to human operators. This pattern recognition capability is foundational for predicting future network demands and potential bottlenecks. Using the insights gained from data analysis, LLMs can forecast periods of high demand across the network, such as during major public events, emergencies, or peak usage times in different geographic areas. With this predictive capability, LLMs can recommend or automate the allocation of resources, including bandwidth, data routes, and power across satellite, aerial, and terrestrial network components.
Moreover, LLMs can automate and enhance decision-making processes in network management by analyzing vast amounts of operational data. This includes predictive maintenance, where LLMs predict equipment failures before they occur, thus minimizing downtime and maintenance costs. In addition, LLMs optimize network traffic flow and resource allocation by processing real-time data from various network segments (satellite, aerial, and terrestrial), ensuring efficient use of network capacity \cite{manias2024towards,guu2020retrieval}.

\subsubsection{Dynamic Traffic Routing} The continuous flow of real-time operational data through ISATNs allows LLMs to make predictions and dynamically adjust resource allocations, computing the most efficient strategies for rerouting data or reallocating resources to ensure the network remains robust and responsive \cite{ren2024tpllm,liu2024spatial}. By anticipating and adjusting to changing network conditions, LLMs help prevent bottlenecks that can lead to congestion and degraded service quality. For instance, if a satellite link is experiencing interference due to atmospheric conditions, an LLM can shift some of the data load to aerial or terrestrial routes before users experience any drop in service quality. To perform these functions effectively, LLMs require initial training on historical data, involving learning from past resource allocation successes and failures. Moreover, the continuous learning abilities of LLM can assist in adapting to new patterns, changes in network usage, and technological advancements to ensure that the resource allocation strategies remain optimal over time \cite{sharma2020toward}.

\subsubsection{Adaptive Bandwidth Management and User Experience} The capabilities of LLMs can assist in adaptive bandwidth management in ISATNs to dynamically allocate and optimize bandwidth in response to real-time network conditions, significantly enhancing user experience. By continuously analyzing historical and live network data, LLMs can predict periods of high demand, adjust bandwidth allocation to prevent congestion, and ensure efficient use of network resources \cite{lee2019survey,naderializadeh2021resource}. This process involves reallocating bandwidth from less critical to more critical areas as network conditions change, thereby maintaining optimal performance and minimizing latency. In addition, LLMs can identify underutilized bandwidth and redistribute it to areas experiencing higher demand, ensuring a balanced and efficient network load. This dynamic adaptability is crucial for maintaining service quality, particularly in high mobility scenarios or varying atmospheric conditions. By optimizing bandwidth allocation, LLMs help to reduce latency, enhance the quality of service for critical applications, and ensure seamless connectivity. Moreover, LLM-driven adaptive bandwidth management enhances user experience by prioritizing bandwidth for high-demand applications such as real-time video streaming and emergency communications, thus ensuring consistent performance even during peak usage times. Accordingly, leading to higher efficiency, better resource utilization, and significantly improved user experience across satellite, aerial, and terrestrial network segments \cite{hussain2020machine}.

\subsubsection{Intelligent Spectrum Management}
LLMs can offer significant potential in enhancing spectrum allocation and sharing within the ISATNs by leveraging their advanced data analysis, prediction, and decision-making capabilities \cite{wang2024llm}. LLMs can identify patterns and predict potential spectrum congestion by processing vast amounts of real-time data from various network components, enabling proactive and dynamic spectrum management \cite{shao2024wirelessllm}. LLMs can use NOMA and CR technologies by providing intelligent decision-making support, such as optimizing NOMA's power levels and predicting CR spectrum availability. This integration of LLMs with advanced spectrum-sharing techniques can significantly improve the overall efficiency and performance of ISATNs.

Furthermore, LLMs can facilitate intelligent spectrum allocation by analyzing historical and real-time data to recommend optimal frequency bands for different network components, considering user demand patterns, geographical information, and network performance metrics \cite{kumar2024ethics}. By dynamically adjusting spectrum allocation based on current conditions, LLMs can help mitigate interference and ensure efficient use of the available spectrum. LLMs can also support collaborative spectrum management across network operators and stakeholders, fostering a more coordinated and efficient spectrum use \cite{qiu2024spectral}.
Despite challenges such as data quality and computational efficiency, integrating LLMs in spectrum allocation and sharing presents a promising approach to addressing the complex requirements of ISATN. This integration ultimately leads to more robust and adaptive communication networks capable of meeting the dynamic demands of modern communication environments.

\subsubsection{Optimizing Dynamic Network Topology}
As LLMs can analyze communication graphs and network traffic data, they can be invaluable for managing dynamic network topologies in ISATNs \cite{lyu2024dynamic}. Additionally, LLMs can generate task-specific code for network management tasks, such as topology modifications and resource allocation, ensuring that the network adapts efficiently to changing conditions. This capability is precious when rapid adjustments are needed to maintain service quality and meet user demands. Integrating LLMs into ISATNs for dynamic topology management also addresses several challenges associated with traditional methods \cite{wurzburg2024NM}. The LLMs can process and understand vast amounts of data, including network performance metrics, topology, telemetry data, and user traffic patterns,  \cite{yin2023joint}. This understanding can then simulate various network scenarios, predict potential bottlenecks or interference issues, and recommend optimal topology configurations to ensure efficient resource utilization and seamless connectivity across the diverse network components in ISATNs. Furthermore, LLMs can facilitate collaboration and coordination among different administrative domains in ISATNs by exposing abstracted views of network topologies and available resources. This collaborative approach enables intelligent decision-making and orchestration of end-to-end network slices spanning multiple domains.

\begin{figure}[h!]
\begin{center}
\includegraphics[width=0.85\columnwidth]{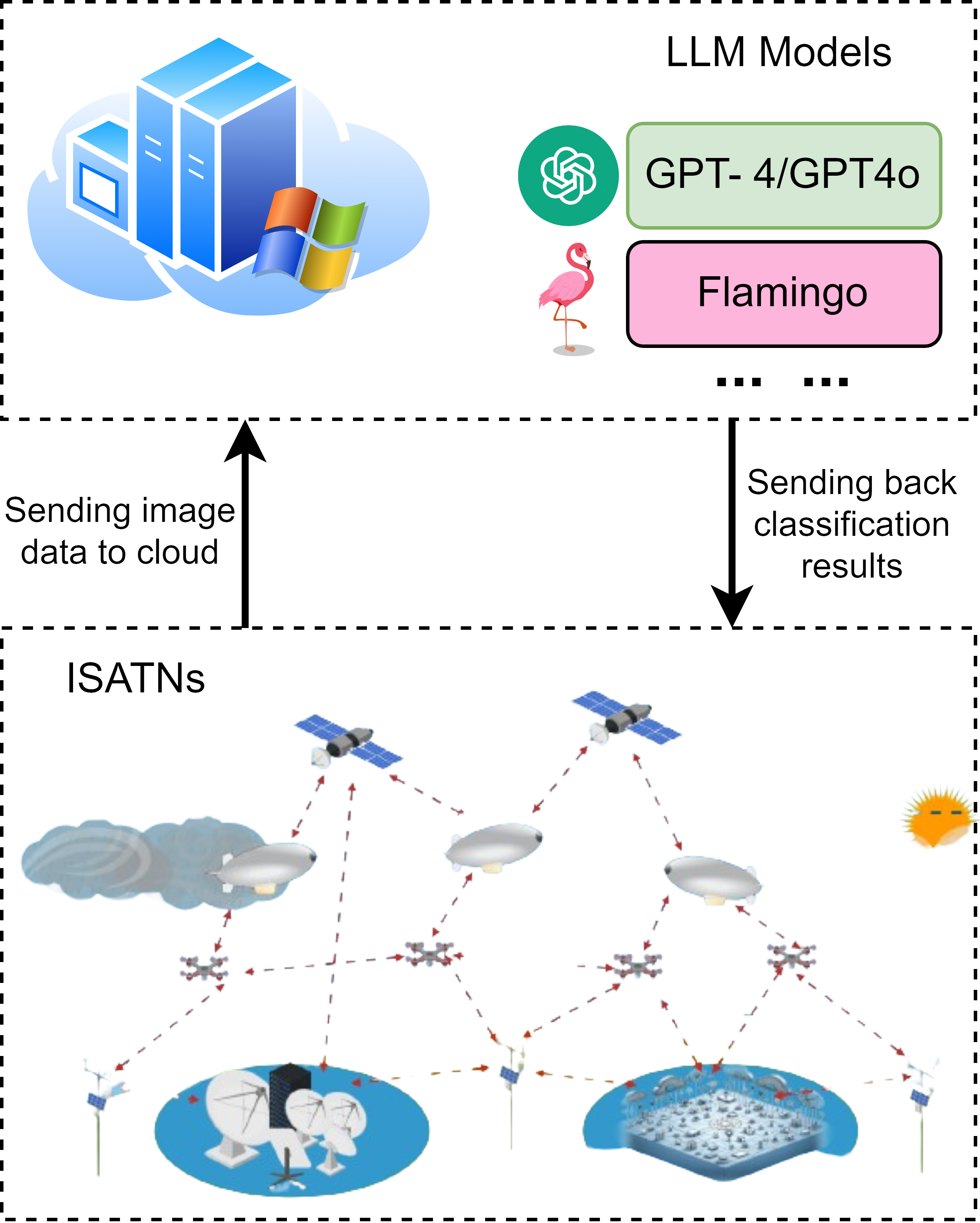}
\caption{Illustration of LLM-assisted spectrum sensing and allocation using computer vision in ISATNs.} \label{compvision}
\end{center}
\end{figure}

Furthermore, the LLMs can automate complex network topology configurations and generate instructions adhering to best practices. This provides real-time monitoring and analysis, allowing proactive measures to address issues promptly. However, integrating LLMs into the ISATN network for dynamic topology optimization depends on the data availability, computational demands, security, trust, and interopeLLMs can use computer vision techniques to collaborative learning, federated training, explainable AI, and adherence to standards will be crucial for realizing the full potential of LLM-enabled dynamic network topology management in ISATNs \cite{lei2021dynamic}. Additionally, techniques like model compression and optimization can enhance LLMs' efficiency and real-time performance in executing time-sensitive network tasks.  For instance, LLM-assisted spectrum sensing using computer vision in ISATNs involves analyzing visual data from spectrum analyzers and environmental sensors to detect and interpret spectrum usage patterns. LLMs can use computer vision techniques to identify interference, detect underutilized frequency bands, and optimize spectrum allocation in real-time. This approach enhances the efficiency and reliability of spectrum management in ISATNs, ensuring optimal performance across satellite, aerial, and terrestrial components. Fig. \ref{compvision} shows an LLM-based solution for spectrum allocation and sharing on a CR network.

\subsubsection{Intelligent Interference Mitigation}
The LLMs can effectively interpret high-level policies and requirements related to interference management in ISATNs \cite{gupta2024llm}. The LLMs can help comprehensively understand the network's state and potential interference sources, facilitating collaboration and coordination among different network entities \cite{Gary2023llm}. LLMs can also address the co-channel interference by training them to predict and recommend optimal power control strategies for ISATN components, such as aerial vehicles and terrestrial base stations, based on real-time interference conditions. By dynamically adjusting transmit power levels, LLMs can mitigate co-channel interference while maintaining adequate signal strength for reliable communication \cite{jacob2024enterprise}. Additionally, LLMs can analyze spectrum usage patterns and recommend optimal channel assignments and frequency planning strategies to minimize co-channel interference across ISATN components. However, integrating LLMs for intelligent interference mitigation in ISATNs poses challenges related to data availability, computational demands, security, trust, and interoperability, which must be addressed through collaborative learning, federated training, explainable AI, and adherence to standards.

 \subsubsection{Latency Minimization and Intelligent synchronization}
LLMs can offer promising solutions to address latency and synchronization challenges in ISATNs by leveraging advanced data analysis, prediction, and decision-making capabilities. By analyzing real-time data and predicting potential delays, LLMs can dynamically adjust network parameters for faster and more reliable communication \cite{proxet2023llm}. LLMs can also enhance synchronization by intelligently coordinating and managing data flows. They can analyze transmission patterns, predict synchronization issues, and implement real-time corrective measures.

Moreover, the LLMs can facilitate the integration of advanced synchronization techniques, such as time-sensitive networking and precision time protocol, into ISATNs. These techniques require precise timing and coordination, which LLMs support by continuously monitoring network performance and adjusting synchronization parameters as needed \cite{nxp2018nxp}. A few key latency metrics, including time to first token, which measures the initial response time, and time to last token, representing the overall processing time \cite{markmunro2024llm}. To minimize latency and improve responsiveness in ISATNs, it is essential to select the appropriate LLM model tailored for the specific use case, optimize the input data length by condensing and prioritizing relevant information, and leverage streaming techniques to start transmitting responses as soon as they are available \cite{googlecloud2023}. Strategies like adjusting the LLM's temperature parameter, setting maximum output token limits, and enabling streaming can enhance perceived responsiveness and create a more interactive and efficient communication experience across the integrated network components.

\subsubsection{Security Management}
LLMs can significantly enhance the security framework of ISATNs by continuously monitoring network traffic across all segments of ISATNs to detect anomalies and potential security threats. Utilizing natural language processing and ML techniques, these models can analyze patterns within vast datasets that might be challenging for traditional systems to process efficiently. LLMs are adept at identifying subtle signs of malicious activity, such as unusual login attempts, spikes in data traffic, or patterns that match known cyber-attack methods \cite{xu2024large}. The capability to analyze and interpret these patterns in real-time allows for a rapid response to potential threats, minimizing the risk of data breaches. In addition, LLMs can automate the generation and revision of security policies by learning from ongoing network activities and evolving security threats. For instance, if an LLM identifies a new distributed denial-of-service attack targeting satellite communications, it can automatically update the firewall rules to mitigate this threat. LLMs can also integrate with IDS and IPS to provide a second layer of analysis to confirm threat detection and recommend or initiate appropriate countermeasures. This rapid response capability is crucial for maintaining network integrity and availability, especially in critical infrastructure systems where downtime or data leakage could have severe consequences \cite{yang2024robustness}. LLMs can analyze past security breaches and incidents to understand how they were executed and how the network responded. This information allows LLMs to learn and adapt, improving their predictive accuracy and the effectiveness of the network's security measures. LLM models can also propose modifications to encryption standards or access controls based on their continuous learning, ensuring that the network's security protocols remain robust against new vulnerabilities \cite{he2024large}.

\newcolumntype{C}{>{\arraybackslash}X} 
\setlength{\extrarowheight}{1pt}
\begin{table*} [htp!]
 \caption{Summary of network management challenges in ISATNs and potential LLMs solutions.}
\label{table22}
 \begin{tabularx}
{\textwidth}{|b|b|b|b|b|}
\hline
\hline
\textbf{Network Management Challenge} & \textbf{Description} &
  \textbf{Potential Solutions Using LLMs} & \textbf{Key Benefits} &
  \textbf{Implementation Considerations}  \\
\hline
Resource Allocation & Allocation network capabilities across terrestrial and non-terrestrial components are complex due to diverse characteristics and dynamic conditions. Requires adaptive strategies and real-time data analysis & LLMs can analyze vast amounts of operational data, predict future network demands, and automate resource allocation. They can also predict periods of high demand and dynamically adjust resource distribution, improving overall network efficiency and performance & Improving resource utilization, reducing costs, and enhancing service quality & Integration with existing network management systems, data privacy concerns \\
\hline
Dynamic Traffic Routing & Efficiently managing network resources, and preventing congestion require advanced analytic tools and adaptive routing protocols. It needs to balance the load and ensure robustness against failures  & LLMs can dynamically adjust resource allocations, compute efficient strategies for rerouting data, and adapt to changing network conditions in real-time. They help prevent bottlenecks and ensure continuous service quality by leveraging predictive capabilities  & Reduced congestion, increased robustness, and constant service availability  &   Scalability, computational overhead \\
\hline
Network Optimization & Enhancing network efficiency and performance involves developing adaptive algorithms and predictive analytics to manage and optimize resources based on real-time data & LLMs can optimize network traffic flow, predict potential network congestion, and adjust resources proactively. They can also enhance load balancing and provide insights for improving network efficiency and reliability & Higher efficiency, proactive problem solving, better user experience & Real-time data integration, and high cost of implementation \\
\hline
Bandwidth Management & Dynamically allocating and optimizing bandwidth to ensure optimal performance, especially under varying network loads and conditions & LLMs can predict bandwidth demands,  identify underutilized bandwidth, and dynamically reallocate resources to areas of higher demand, ensuring balanced and efficient network load & Improved bandwidth utilization, reduced latency, better quality of service & Monitoring tools for real-time adjustments, integration with existing systems \\
\hline
\end{tabularx}
\end{table*}

\begin{table*} [htp!]
 \begin{tabularx}
{\textwidth}{|b|b|b|b|b|}
\hline
Scalability Issues & Scaling up the network to handle increased loads without compromising performance and reliability & LLMs can help forecast network load trends and scale resources accordingly. They can also optimize the distribution of resources across a more extensive network to maintain performance & Enhanced scalability, efficient resource use, consistent performance & Infrastructure upgrades, cost considerations \\
\hline
Latency Reduction & Minimizing latency for real-time applications such as video conferencing, online gaming, and autonomous vehicles & LLMs can predict and mitigate latency issues by optimizing data paths and reducing transmission delays through intelligent routing and real-time adjustments & Lower latency, improved real-time performance, better user satisfaction & Network infrastructure enhancements, monitoring and adjustments \\
\hline
Spectrum Allocation and Sharing & Optimizing the use of available frequency bands and ensuring efficient communication across satellite, aerial, and terrestrial components, each with unique spectrum requirements and dynamic conditions & LLMs can analyze spectrum usage patterns, predict interference scenarios, and suggest optimal allocation strategies. They can enable dynamic spectrum access and coordinate spectrum sharing among diverse network components & Improve spectrum efficiency, reduce interference, and enhance overall network performance & Integration with existing spectrum management frameworks, ensuring regulatory compliance and maintaining spectrum sensing accuracy \\
\hline
Dynamic Network Topology Management & Adapting to rapidly changing network topologies in real-time to maintain efficient and reliable communication across ISATN components, considering factors like user mobility and varying traffic loads & LLMs can continuously monitor and analyze network parameters, predict changes, and dynamically adjust the network topology. They can optimize routing paths and resource allocation to adapt to evolving conditions & Enhance network resilience, reduce latency, and improve overall performance & Real-time data integration, scalability of topology management algorithms, and handling computational overhead \\
\hline
Interference Management & Managing various types of interference (SI, MI, etc.) across satellite, aerial, and terrestrial components to ensure seamless communication & LLMs can detect and predict interference patterns, suggest mitigation strategies, and optimize resource allocation to minimize interference. They can also adapt to changing conditions and provide real-time solutions & Enhanced signal quality, reduced bit error rates, improved network capacity & Regular updates to interference models, integration with interference suppression and avoidance techniques, and coordination among network components \\
\hline
\end{tabularx}
\end{table*}

\begin{table*} [htp!]
 \begin{tabularx}
{\textwidth}{|b|b|b|b|b|}
\hline
Latency and Synchronization & Addressing the latency and synchronization issues arising from the diverse nature of ISATN components and ensuring seamless data transmission across vast distances, particularly with satellite links & LLMs, can predict latency issues, optimize data paths, and adjust synchronization mechanisms in real-time. They can ensure coordinated data transmission across different network segments with varying latency characteristics  & Lower latency, improved synchronization, and better QoS for delay-sensitive applications & Enhance network infrastructure, maintain synchronization accuracy, and integrate with existing timing references and protocols \\
\hline
Security Management & Ensuring robust security protocols across diverse network environments is challenging due to varying operational and regulatory standards. Requires continuous monitoring and rapid response to threats & LLMs continuously monitor network traffic, detect anomalies, and automate security updates. They enhance predictive accuracy and network resilience & Enhanced security, rapid threat mitigation, improved compliance & Regular updates to threat models, and compliance with regulations \\
\hline
\hline
\end{tabularx}
\end{table*}

\section{Technical LLMs Considerations in ISATNs}\label{sec:04}
\begin{figure}[h!]
\begin{center}
\includegraphics[width=0.75\columnwidth]{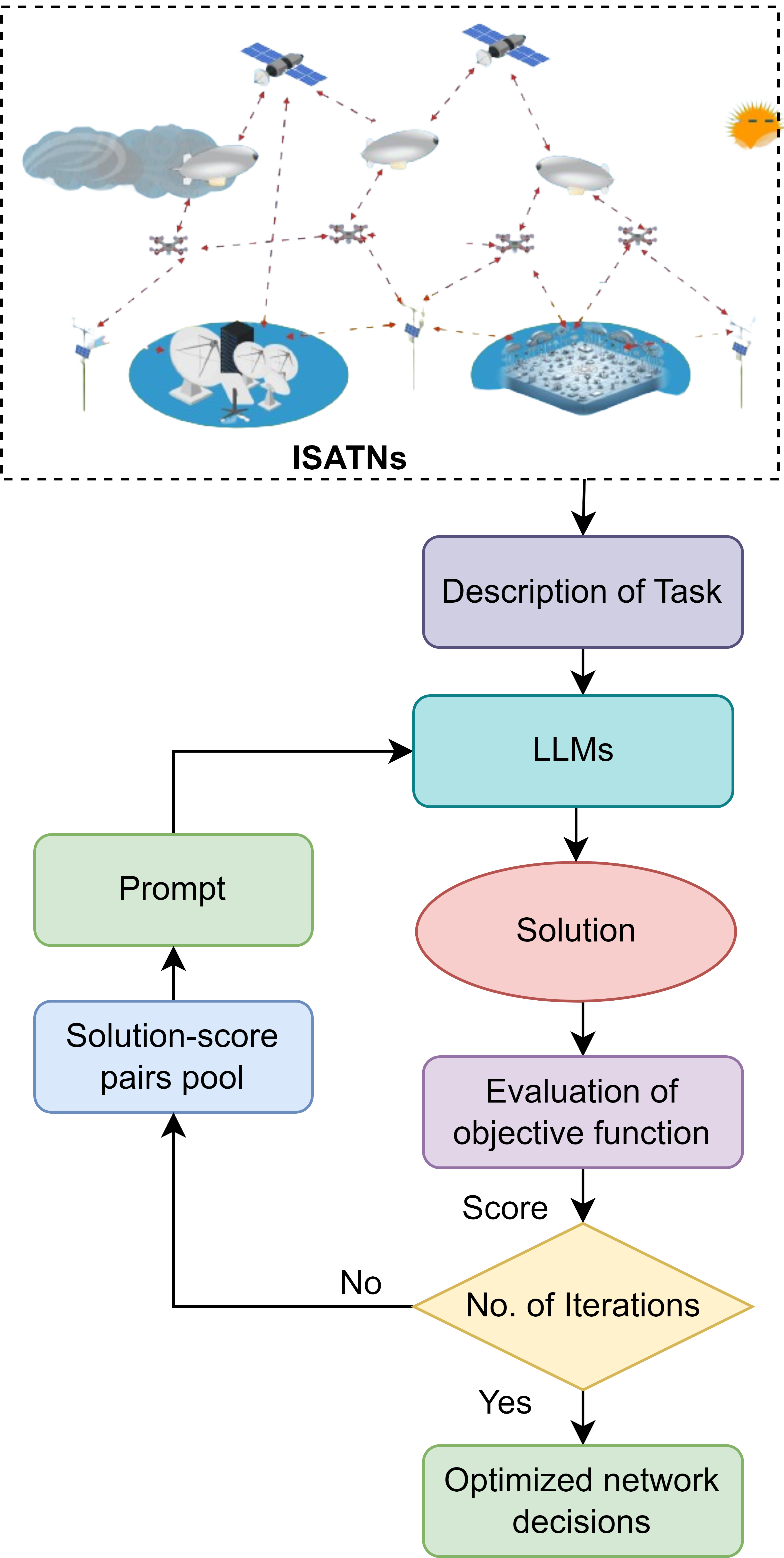}
\caption{Illustration of LLM as a network optimizer for ISATNs.} \label{optimizationofISATNs}
\end{center}
\end{figure}

Although LLMs can significantly benefit ISATNs, they also face several technical challenges. Key considerations include data integration difficulties stemming from heterogeneous network components, effective data preprocessing and compatibility strategies, and the curation of specialized datasets. Additionally, we examine efficient training techniques and underscore the importance of large datasets and specialized training procedures to achieve optimal LLM performance within the ISATN domain. For example, LLMs can function as optimizers for ISATNs by leveraging their advanced data processing and predictive capabilities to dynamically manage and optimize network resources. Fig. \ref{optimizationofISATNs} shows the generic illustration of LLMs used as optimizers for the ISATNs. Analyzing the real-time data from satellite, aerial, and terrestrial components, LLMs can predict network conditions, optimize resource allocation, and enhance overall network performance without requiring detailed knowledge of their internal workings. This approach allows for efficient handling of the complex and dynamic nature of ISATNs, improving connectivity, reducing latency, and ensuring seamless communication across the network. In the following, we discuss these challenges in detail.

\subsection{Data Integration and Heterogeneity}
Integrating data from diverse network components is crucial for LLM-based ISATNs to comprehensively understand the network's state and make informed decisions regarding interference management, resource allocation, and topology optimization \cite{zhou2024llm, bai2011stack}. However, this process faces challenges such as data format compatibility, synchronization, and scalability due to the heterogeneous nature of data sources, including aerial vehicles and terrestrial networks, each with different data formats, structures, and protocols. For instance, an exciting experiment in \cite{dominik2023llm} demonstrates how LLMs can merge different data formats without human intervention \cite{dominik2023llm}. Integrating JSON-like data with an HTML table from a website highlights the challenge of heterogeneous data formats. Traditionally, this integration has been time-consuming and costly, often leading to underutilized data and missed opportunities for insights \cite{xu2024cached}. Effective data integration strategies are essential to address these challenges.

LLMs can offer a promising solution by leveraging their natural language processing capabilities to understand and interpret diverse data formats, ranging from structured formats like JSON and XML to unstructured formats like HTML tables or proprietary data structures used by different network components \cite{xu2024cached, googlecloud2023}. LLMs can be prompted to extract relevant data from these sources, parse it into a standard format, and seamlessly integrate it with data from other components, enabling a unified view of the network's state and performance. However, integrating LLMs for data integration in ISATNs also poses challenges, such as ensuring data privacy and security, as sending raw data to third-party LLM providers may violate data protection policies \cite{dominik2023llm}. A more secure approach to mitigate this is to generate code using LLMs, where Python functions or scripts can perform data transformation and integration on-premises without exposing sensitive data\cite{luca2024llm}. Furthermore, computational demands, reliability, and interoperability challenges must be addressed through collaborative learning, federated training, explainable AI, and adherence to standards.

Preprocessing techniques are essential for transforming raw data into a format compatible with LLMs for effective processing and analysis in ISATNs. These techniques include data cleaning and normalization, feature engineering, and multimodal data integration and alignment \cite{xu2024fairly}. Data cleaning and normalization involve identifying and removing irrelevant or redundant information, handling missing values, and standardizing data formats across different sources \cite{hua2023index}. Feature engineering involves transforming raw data into meaningful features that LLMs can effectively utilize, such as extracting relevant information from satellite imagery, aerial sensor data, and terrestrial sensor networks using techniques like image segmentation, object detection, and signal processing \cite{ye2024language}. For multimodal data, which combines different data types such as text, images, audio, and video, preprocessing techniques align these modalities. Techniques like data fusion and multimodal representation learning can combine and represent multimodal data in a format that LLMs can effectively process. Transfer learning techniques can also leverage pre-trained models on large datasets, enabling efficient processing of multimodal data in ISATNs. Despite the challenges, the potential benefits of LLM-enabled data integration and preprocessing in ISATNs, such as improved decision-making, resource optimization, and enhanced network performance, make it a promising area for further exploration and development.

\subsection{Training LLMs for ISATN Tasks}
Training LLMs for ISATN-specific tasks requires a comprehensive approach to address the unique challenges posed by the diverse and multimodal nature of data in these networks. A crucial aspect is curating and preparing high-quality training data, which involves collecting and integrating data from sources like satellite imagery, aerial sensors, and terrestrial sensor networks \cite{hoffmann2022training}. Data augmentation, transfer learning, and domain adaptation can enhance the training data's quality and diversity, enabling LLMs to generalize better to real-world ISATN scenarios. Preprocessing techniques like data cleaning, normalization, and feature engineering are essential to transform raw data into a format compatible with LLMs.

Selecting and fine-tuning appropriate LLM architectures and pre-trained models is another critical consideration. Large-scale pre-trained LLMs like GPT and BERT provide a strong foundation, but fine-tuning these models on ISATN-specific data is crucial for optimal performance \cite{biderman2023pythia}. Techniques such as multi-task learning, where the LLM is trained on multiple related tasks simultaneously, and curriculum learning, which exposes the model to increasingly complex tasks in a structured manner, can be beneficial. SmoothQuant offers a training-free, accuracy-preserving post-training quantization solution that enables 8-bit weight and activation quantization for LLMs, achieving significant speedup and memory reduction with negligible accuracy loss, thus reducing hardware costs and democratizing LLMs \cite{carlini2021extracting}.

Given the distributed nature of ISATNs, leveraging distributed and federated learning approaches is essential. Distributed training techniques allow for parallel data processing across multiple nodes, facilitating efficient training on large-scale datasets \cite{kandpal2023large}. Federated learning enables collaborative training of LLMs while preserving data privacy and security, as the training data remains decentralized across different locations. As data and operational conditions evolve, continuous learning and adaptation are also crucial for LLMs in ISATNs. Techniques like online, incremental, and meta-learning can enable LLMs to adapt and improve continuously as new data becomes available or operational conditions change.

By addressing these challenges and leveraging advanced training techniques, LLMs can be effectively trained to tackle various tasks in ISATNs, such as satellite image analysis, aerial surveillance, terrestrial sensor data processing, and multimodal data fusion \cite{xiao2023smoothquant}. Training LLMs for ISATN tasks will ultimately empower these integrated networks, enabling more efficient and effective decision-making, resource allocation, and situational awareness.

\subsection{Large Datasets Requirement}
LLMs have shown exceptional capabilities in various natural language processing tasks, such as text generation, summarization, and question-answering. However, their performance largely depends on the quality and quantity of training data and the fine-tuning techniques used \cite{muennighoff2024scaling,yu2024large}. A balance between large datasets and specialized training is crucial in ISATNs, where data from multiple sources and modalities must be processed. One significant advantage of LLMs is their ability to recognize and utilize patterns and relationships within vast datasets \cite{muennighoff2024scaling}. LLMs can develop a deep understanding of language, context, and domain-specific knowledge by training on extensive, diverse datasets.

For example, in \cite{telmai2023max}, the authors used the 'blbooksgenre' dataset to classify digitized books into Fiction and Non-Fiction categories using an LLM. By introducing noise to the data to mimic real-world scenarios, they used DistilBert, a transformer-based LLM, to study performance fluctuations when trained on datasets with errors \cite{shi2023detecting}. For ISATN tasks, large datasets that include satellite imagery, aerial data, terrestrial sensor data, and related textual information are needed. These datasets enable LLMs to learn complex relationships between different data sources, leading to more accurate and comprehensive analysis and decision-making.

Specialized training techniques tailored to ISATN tasks can significantly enhance LLM performance. This may involve fine-tuning models on domain-specific datasets, using multi-modal data fusion techniques, or applying transfer learning approaches. For instance, LLMs can be pre-trained on general-purpose datasets and then fine-tuned on ISATN-specific data to adapt to the unique challenges of this domain. Multi-task and curriculum learning can also improve the models' ability to handle diverse tasks and data modalities encountered in ISATN scenarios.

ISATN environments are dynamic and constantly evolving, with new data sources, modalities, and challenges. Therefore, LLMs deployed in ISATN systems must be capable of continuous learning and adaptation. Techniques such as online learning, where models are periodically updated with new data, and transfer learning approaches can help LLMs quickly adapt to new tasks or domains within the ISATN ecosystem \cite{shi2024continual}. The success of LLMs in ISATN tasks relies on large, diverse, high-quality datasets and specialized training techniques tailored to this domain's unique challenges. By leveraging these resources and methodologies, LLMs can unlock their full potential in empowering ISATN systems, enabling more efficient data analysis, decision-making, and overall system performance \cite{li2024examining}.

\subsection{Resource Constraints}
The high computational requirements for training and running LLMs demand substantial hardware resources, which can be challenging in distributed ISATN environments. Managing the energy consumption of deploying LLMs is crucial, particularly for satellite and aerial platforms with limited power supplies. LLMs are resource-intensive, posing significant challenges regarding computational power, memory, and data storage \cite{dhar2024empirical, oh2024exegpt}.
Since ISATNs involve diverse data sources and dynamic wireless channel conditions, LLMs are required to efficiently process and analyze large volumes of data in real time. This necessitates significant computational resources, often limited and distributed across multiple components in ISATNs \cite{doshi2023llm}. Future research should focus on developing LLMs that operate efficiently within these constraints. This includes optimizing LLM architectures to reduce computational overhead, leveraging edge computing and distributed processing to minimize latency, and developing specialized hardware for hybrid network conditions \cite{xu2024survey}.
Future work should also explore techniques for compressing and processing large data volumes to reduce the computational burden on LLMs. This can involve data compression, feature engineering, and multimodal data integration to ensure efficient processing and analysis \cite{liu2024enhancing}.

\section{Future Research Directions}\label{sec:05}
In this section, we discuss the promising future research directions for utilizing LLMs in ISATN. Potential areas include developing multimodal LLMs capable of seamlessly integrating diverse data integration, latency, effective decision-making, terahertz communication, effective power communication, and secure LLM architectures for responsible deployment in critical ISATN applications.

\subsection{Specialized LLMs for Wireless Channel Modelling}
Understanding and processing real-time data from wireless channels necessitates that LLMs interpret complex, dynamic information about signal strength, interference, noise levels, and bandwidth fluctuations. The significant environmental variability due to weather, terrain, and atmospheric changes, coupled with user mobility—especially in terrestrial and aerial segments—and physical obstructions such as buildings, mountains, or vegetation, can impact the reliability and strength of connections \cite{wei2021hybrid,zhu2021integrated}. Consequently, LLMs must be highly adaptive and capable of learning from diverse scenarios to predict and respond to network conditions. This adaptability helps mitigate signal propagation issues and maintains high connectivity and service quality. Moreover, given the multiple data sources and continuous transmission in ISATNs, managing this data flow effectively without bottlenecks necessitates highly optimized algorithms and robust computational resources. Therefore, future integrated LLM-based systems must efficiently process the high volumes of data generated by ISATNs, ensuring fast and consistently reliable performance that does not degrade over time or under stress.

In addition, data collected from satellite, aerial, and terrestrial components in ISATNs vary significantly in type, frequency, and format, impacting real-time wireless channel information. For instance, satellite data is generally broad and less frequent, whereas terrestrial sensors often provide high-frequency updates. Future work should focus on unifying sophisticated data fusion techniques to accurately align, correlate, and synthesize this diverse information to enhance wireless channel management's real-time responsiveness and effectiveness. Research should also focus on applying transfer learning to adapt models pre-trained on large datasets to the specific wireless communication scenarios encountered in ISATNs. This approach would enhance the efficiency and applicability of LLMs for addressing the unique challenges associated with integrating satellite, aerial, and terrestrial networks \cite{wang2021transfer,liu2019machine,nguyen2022transfer}.

Moreover, to further improve real-time wireless channel management using LLMs, novel approaches should focus on developing customized model architectures to handle the challenges of high variability and real-time wireless communication data. This includes effectively managing temporal and spatial data and integrating adaptive learning techniques for continuous system improvement. Future efforts should also explore leveraging edge computing to reduce latency and provide real-time insights into the wireless channel's status \cite{qiu2022mobile,zhang2019satellite,xie2020satellite,yu20203d}.

\subsection{Advancing Data Integration Techniques}
Although significant progress has been made in developing data integration techniques for ISATNs, the potential of LLMs remains largely unexplored in this context. Therefore, future work should focus on developing adaptive data fusion techniques that can dynamically integrate and analyze data from disparate sources within ISATNs. Integrating LLMs can enhance the data processing capabilities for vast amounts of heterogeneous data, thereby providing a cohesive view of the network state. Novel network management schemes should also focus on employing LLMs to enhance network management capabilities in ISATNs, such as dynamic bandwidth allocation, load balancing, and fault diagnosis. These models could utilize integrated data to predict network behavior and adjust real-time configurations based on changing channel characteristics \cite{xue2024wdmoe,jiang2024personalized}.

Future work should also explore semantic data integration methods that leverage LLM capabilities to understand the context and meaning behind the data collected from ISATNs to resolve ambiguities and enhance the accuracy of data interpretation across the network. Future research should advance data integration techniques by developing cross-layer optimization strategies in dynamic ISATNs using LLMs to enhance multiple network layers (i.e., from the physical to the application layer) to improve overall network efficiency, reliability, and performance. Researchers should also focus on implementing cognitive networking capabilities using LLMs to enhance the adaptability of ISATNs to dynamic wireless channel characteristics and user demands. These capabilities will enable the network to self-learn and dynamically adjust its operations through continuous data integration and analysis \cite{mountantonakis2019large,de2018using,chen2019efficient}.

\subsection{Low-Latency and Low-Complexity Solutions}
The research focuses on reducing the latency and computational demands of LLMs in real-time applications, which is particularly pertinent in ISATNs. The complex dynamics between satellite and terrestrial components in these networks present unique challenges that necessitate minimizing latency to ensure seamless communication services, such as real-time voice recognition and autonomous vehicle coordination. Future work should focus on integrating edge computing to enable LLMs near the edge of satellite and terrestrial networks to reduce transmission delays \cite{shen2024large,lin2023pushing,chen2019deep}. Additionally, network slicing should be explored to optimize bandwidth allocation within these hybrid networks, prioritizing critical LLM traffic. To address the high computational overhead associated with LLMs, research should focus on model optimization techniques such as pruning, quantization, and knowledge distillation, which can enhance the performance of ISATNs while reducing the computational burden \cite{dandoush2024large,khan2020network}. Furthermore, developing specialized hardware tailored for hybrid network conditions is crucial in managing the computational demands of ISATNs effectively. The research efforts should leverage LLMs' advanced processing and analytical capabilities to tackle the unique challenges posed by integrating satellite and terrestrial network data, particularly in dynamically changing environments \cite{dhar2024empirical,zeng2024implementation}.

\subsection{Explainable and Trustworthy LLMs-Based Network Decisions}
With LLMs becoming increasingly integral to network management in dynamic and complex wireless communication environments, building systems that operators and stakeholders can understand and trust is crucial. Future research should prioritize the development of frameworks that ensure the explainability and trustworthiness of LLM-based decisions within ISATNs. Novel methods should focus on developing algorithms and tools that can articulate the reasoning behind LLM-driven decisions, making it easier for network administrators to interpret and validate these decisions using visualizations, user-friendly interfaces, and detailed logging mechanisms that capture the decision-making process of LLMs in real-time. Future work should explore robust validation and verification techniques to ensure that LLM-based decisions are accurate, consistent, and secure. This includes developing frameworks for continuous monitoring and auditing of LLM outputs, implementing robust security measures to protect against adversarial attacks, and establishing protocols for the human oversight of automated decisions \cite{mallen2022not,kim2024m,chang2024survey}.

\subsection{Intelligent Reflecting Surfaces (IRS) for LLMs-assisted ISATNs}
IRS technology has the potential to significantly enhance wireless communication by dynamically managing signal propagation, thereby improving coverage, capacity, and overall network performance \cite{kong2023uplink,xu2021intelligent}. Future research should benefit from the broad advantages of LLM in integrating IRS within ISATNs to optimize the placement and control of IRS elements. Future schemes should use LLM to analyze vast network data and environmental variables to determine the optimal placement of IRS elements to provide real-time recommendations for adjusting IRS configurations to maximize signal strength and quality. Novel reinforcement learning algorithms should allow LLMs to continuously learn from network conditions and user interactions, enabling adaptive control of IRS elements. It involves creating models to predict environmental changes and proactively adjusting IRS settings to maintain optimal communication links. Additionally, future work should address the computational challenges associated with real-time IRS control, exploring how edge computing and distributed processing can support the deployment of LLMs in managing IRS \cite{chu2020intelligent,bai2020latency,sun2022energy}. Ensuring the explainability and trustworthiness of LLM-driven decisions in IRS control is also crucial, necessitating the development of transparent frameworks that operators can understand and trust.

\subsection{LLMs for THz-based ISATNs}
State-of-the-art literature has paid significant attention to understanding the potential of THz communication, particularly its ability to provide ultra-high bandwidth for modern communication systems. Current studies have explored basic propagation models, initial resource allocation strategies, and early integration efforts with existing network infrastructures \cite{xing2021terahertz,nie2021channel,wang2023coverage}. However, these efforts have addressed fundamental challenges without leveraging advanced ML techniques like LLMs. Given the complexity and potential of THz communication and the advanced capabilities of LLMs, it is crucial to focus on integrating these technologies to enhance the efficiency and effectiveness of ISATN systems. Thus, future research should explore the potential of LLMs for resource allocation and channel modeling within the THz spectrum to offer ultra-high bandwidth, which is essential for meeting the increasing data demands of ISATNs \cite{duong2023machine,puspitasari2023emerging}. It also necessitates addressing the unique propagation challenges, such as high atmospheric absorption and limited transmission distance. Future work should focus on leveraging the advanced processing capabilities of LLMs to develop sophisticated models that accurately predict channel behavior and propagation characteristics to enable dynamic adjustment of network parameters to maintain optimal performance and reliability. Future algorithms should also focus on creating algorithms that allow LLMs to learn from real-time network conditions and user interactions, facilitating adaptive resource management and channel modeling. These algorithms could help mitigate the adverse effects of THz propagation challenges by proactively adjusting transmission strategies based on current environmental conditions. Additionally, integrating edge computing with LLMs could further enhance the efficiency and responsiveness of THz communication in ISATNs by processing data closer to the source to reduce latency and support real-time decision-making for resource allocation and channel adjustments \cite{zhu2021integrated}.

\subsection{Dynamic Spectrum Access}
One of the primary challenges in ISATNs is efficiently managing the spectrum, given its dynamic nature and the need to accommodate varying demands from both satellite and terrestrial components. Traditional static spectrum allocation methods often lead to suboptimal usage and increased interference \cite{jia2020intelligent,liu2021spectrum}. Future work should focus on integrating LLMs with deep RL techniques to continuously learn and adapt to the spectrum environment in real-time, making proactive decisions about spectrum access and allocation. This dynamic approach can significantly reduce interference, improve bandwidth utilization, and ensure higher quality of service for end-users. Also, the studies should focus on designing Reinforcement Learning (RL) algorithms to enable ISATNs to learn from past experiences and adjust strategies based on real-time data. When combined with the predictive capabilities of LLMs, these algorithms can foresee spectrum usage patterns and environmental changes, allowing for more informed and effective decision-making processes.

\begin{figure}[h!]
\begin{center}
\includegraphics[width=0.9\columnwidth]{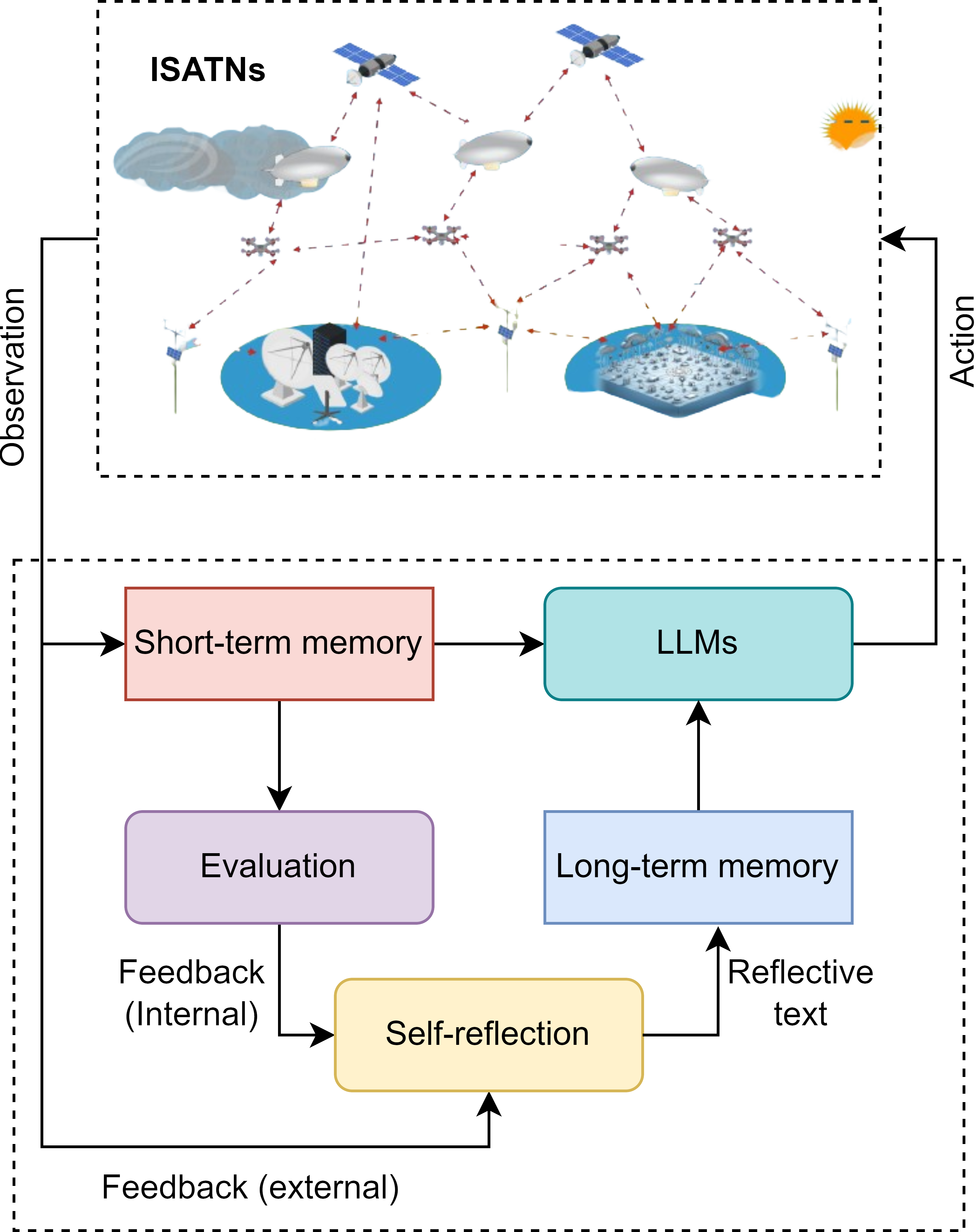}
\caption{Illustration of LLM-assisted dynamic spectrum access in ISATNs.} \label{RL_ISATNs}
\end{center}
\end{figure}

Moreover, LLMs can process large volumes of heterogeneous data from various network sources to identify patterns and insights that are not immediately apparent \cite{dong2023lambo}. RL can then use these insights to develop strategies for optimal resource allocation, ensuring that spectrum is used most efficiently by balancing factors such as interference levels, user demand, and signal strength. Fig. \ref{RL_ISATNs} depicts an example of an LLM-assisted RL-based dynamic strategy for spectrum access in ISATNs. Based on current needs and predicted usage, this network can dynamically shift resources between satellite and terrestrial networks. Future work should focus more on real-time adaptability by integrating LLMs and RL, allowing ISATNs to respond swiftly to changes in the spectrum environment. This real-time adaptability is crucial for maintaining optimal network performance during unexpected events, such as sudden surges in user demand or interference from other networks. The ability to quickly adapt and reconfigure the network will lead to continuous, high-quality service and minimized downtime \cite{jiang2019language,dai2019big}.

\subsection{Power Efficient Communication}
Existing studies have extensively explored various power-saving techniques, including energy-efficient routing, dynamic power adjustment, energy-efficient beamforming, and integrating renewable energy sources \cite{alagoz2011energy,huang2020energy,ruan2017energy,hassan2022seamless}. However, these efforts have often relied on conventional methods and have not fully leveraged the advanced capabilities of LLMs. Future research should investigate LLMs to develop adaptive power management strategies for wireless communication within ISATNs to balance network performance with energy consumption, ensuring efficient operation while maintaining high-quality service. The primary focus should be on creating adaptive power management algorithms that utilize the predictive capabilities of LLMs. These models can analyze historical and real-time data to forecast network traffic patterns and environmental conditions, allowing the network to adjust power levels dynamically. By predicting high and low-demand periods, the network can allocate power more efficiently, reducing energy consumption during off-peak times without compromising performance \cite{manias2024towards,dandoush2024large}.

Future work should leverage LLMs training capabilities to predict energy consumption based on various factors such as user behavior, network load, and environmental conditions \cite{dai2019big,kibria2018big,zeng2024implementation}. It is imperative to explore the integration of these predictive models into network management systems to provide real-time insights into energy usage. This will enable proactive adjustments to power settings, leading to more sustainable energy use across the network. Future energy optimization algorithms should consider adjusting transmission power, modulating signal strength, and optimizing the deployment of network resources based on current and predicted demands. Developing energy-efficient communication protocols guided by LLM insights should be a key area of research to maximize data transmission with reduced unnecessary energy expenditure. For example, adaptive modulation and coding schemes can lower power usage during stable conditions, while higher power levels can be reserved for more challenging communication scenarios. Future work should also consider how LLMs can support the integration of renewable energy sources into ISATNs. By predicting energy availability from sources like solar and wind, LLMs can help manage the network's energy mix, optimizing the use of renewable energy while ensuring consistent performance. This integration can further enhance the sustainability of ISATNs. Future work should also explore the distributed processing mechanisms for LLM predictions and adjustments locally to reduce the need for data to travel to centralized servers to minimize latency and improve the efficiency of power management strategies \cite{shen2024large,lin2023pushing,yu2024edge,chen2019deep}.

\section{Conclusion} \label{sec:06}
This survey comprehensively examines the integration of LLMs into ISATNs by providing a detailed analysis of how various components of ISATNs (such as satellite communication layers, aerial drone swarms, and ground-based network systems) can benefit from LLM integration. The study highlights that LLMs can significantly reduce latency, optimize data flow, improve signal processing, and enable more effective network traffic management through advanced predictive algorithms and real-time decision-making capabilities. By exploring various use cases, the paper further demonstrates how LLM-based ISATNs can enhance the performance of existing network systems. However, it also acknowledges the technical limitations of integrating LLMs into ISATNs, such as data integration challenges, scalability issues, and data computation demands. To address these limitations and fully exploit the potential of LLMs in ISATNs, this work identifies several future research directions. The future direction focuses on integrating LLMs to enhance the integration of intelligent reflecting surfaces, exploiting THz communication, dynamically managing spectrum access with deep reinforcement learning, and implementing energy-efficient strategies, all within the framework of adaptive and robust network operations in ISATNs. In conclusion, this study underscores the transformative potential of LLMs for ISATNs in enhancing operational efficiency, advancing 5G/6G communication technologies, and overcoming traditional data transmission and processing bottlenecks. By addressing current challenges and future research directions, this work sets the stage for continued innovation and improvement in network technologies, ultimately contributing to advancing ISATNs and global communication networks.

\bibliographystyle{IEEEtran}
\bibliography{draft}

\end{document}